\documentclass[11pt]{article}
\PassOptionsToPackage{table}{xcolor}
\usepackage[preprint]{acl} 
\usepackage{colortbl}
\usepackage{times}
\usepackage{latexsym}
\usepackage[T1]{fontenc}
\usepackage[utf8]{inputenc}
\usepackage{microtype}
\usepackage{inconsolata}
\usepackage{url}
\usepackage{booktabs}
\usepackage{graphicx}
\usepackage{xcolor}
\usepackage{array}
\usepackage{tikz}
\usetikzlibrary{patterns}
\usepackage{listings}
\usepackage{multirow}
\usepackage{tabularx}
\usepackage{rotating}
\usepackage{makecell}
\usepackage{enumitem}
\usepackage[most]{tcolorbox}
\usepackage{ragged2e}
\usepackage{placeins}
\usepackage{fvextra}
\definecolor{cedarred}{RGB}{165,45,45}
\definecolor{cedargreen}{RGB}{20,120,70}

\newtcolorbox{metricbox*}[1]{
  enhanced,
  breakable,
  colback=gray!5!white,
  colframe=gray!40!black,
  fonttitle=\bfseries,
  title=#1,
  boxsep=2pt,
  left=4pt, right=4pt, top=3pt, bottom=3pt,
  arc=2mm,
  toptitle=1mm, bottomtitle=1mm,
}

\usepackage{amsmath,amsfonts,bm}

\def\eqref#1{equation~\ref{#1}}

\def\1{\bm{1}}

\DeclareMathAlphabet{\mathsfit}{\encodingdefault}{\sfdefault}{m}{sl}
\SetMathAlphabet{\mathsfit}{bold}{\encodingdefault}{\sfdefault}{bx}{n}

\newcommand{\papertitle}{CEDAR-GRPO: Process-Aware Reinforcement Learning for General Abductive Reasoning in LLMs}

\newcommand{\titleicon}[1][0.9em]{%
  \raisebox{-0.15\height}{\includegraphics[height=#1]{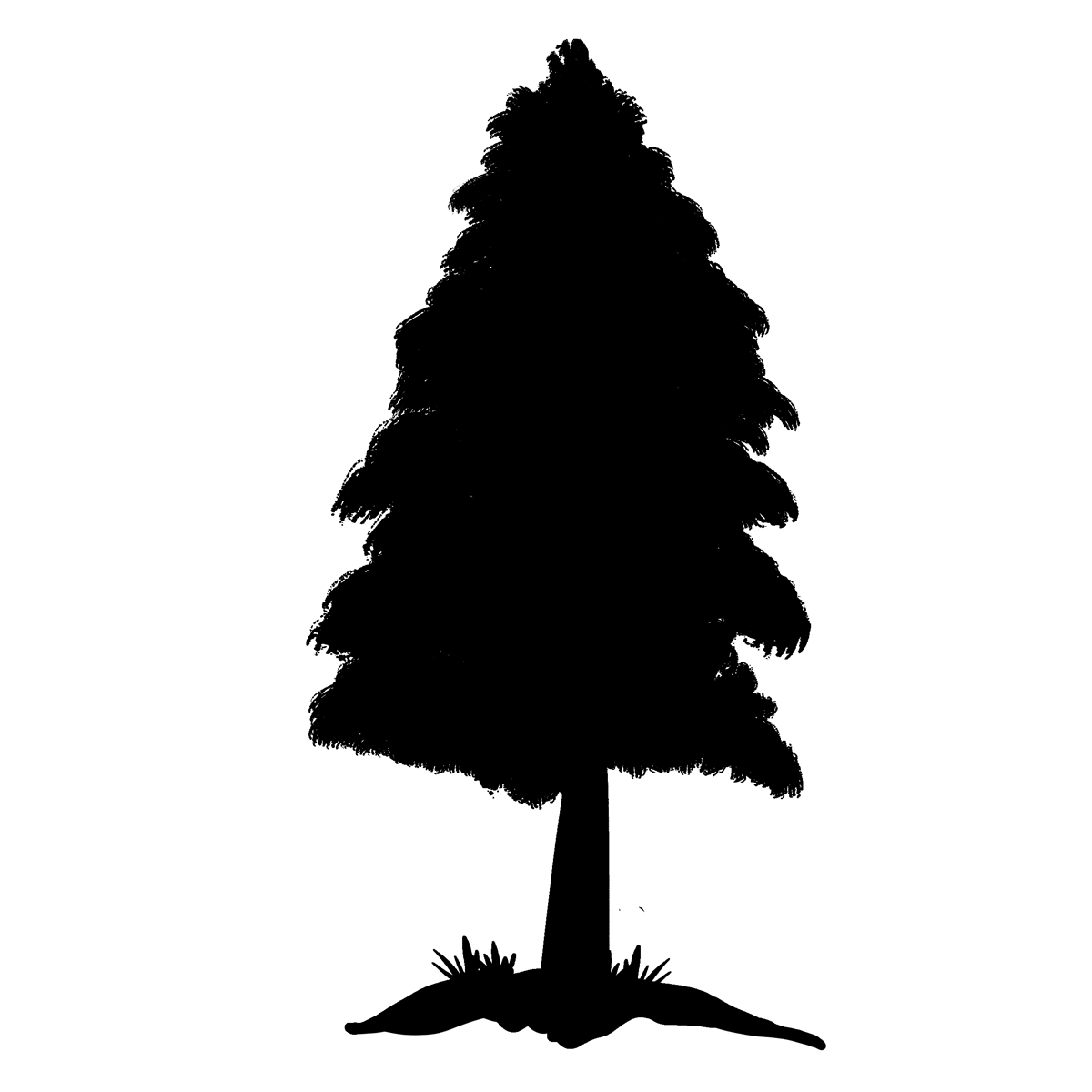}}%
}

\title{\papertitle}

\makeatletter
\AtBeginDocument{%
  \@ifpackageloaded{hyperref}{%
    \hypersetup{pdftitle={CEDAR-GRPO: Process-Aware Reinforcement Learning for General Abductive Reasoning in LLMs}}%
  }{}%
  \let\CEDAR@orig@maketitle\@maketitle
  \def\@maketitle{%
    \begingroup
      \def\@title{\titleicon[0.95em]\hspace{0.35em}\papertitle}%
      \CEDAR@orig@maketitle
    \endgroup
  }%
}
\makeatother

\author{%
\normalfont
\begin{tabular}{c}
\textbf{Moein Salimi}\textsuperscript{1} \quad
\textbf{Danial Parnian}\textsuperscript{1} \quad
\textbf{Shaygan Adim}\textsuperscript{1} \\
\textbf{Amirmohammad Ebrahiminasab}\textsuperscript{2,*} \quad
\textbf{Nima Alighardashi}\textsuperscript{1,*} \quad
\textbf{Parsa Gholami}\textsuperscript{1,*} \\
\textbf{Sahand Akramipour}\textsuperscript{1} \quad
\textbf{Mahdi Jafari Siavoshani}\textsuperscript{1,\textdagger} \quad
\textbf{Mohammad Hossein Rohban}\textsuperscript{1,\textdagger} \\[0.35em]
{\normalsize \textsuperscript{1}\,Sharif University of Technology \quad
\textsuperscript{2}\,University of Tehran} \\
{\normalsize \texttt{\{mjafari,rohban\}@sharif.edu}}
\end{tabular}%
}

\begin{document}

\maketitle

\begingroup
\renewcommand{\thefootnote}{\fnsymbol{footnote}}
\footnotetext[1]{Equal contribution.}
\footnotetext[2]{Equal corresponding authors.}
\endgroup

\begin{abstract}
Abductive reasoning, often characterized as inference to the best explanation, is central to explanation under uncertainty, from everyday sense-making and investigation to scientific discovery. Yet LLM research has mostly studied abduction through narrow, task-specific benchmarks, making it unclear whether observed gains transfer beyond the benchmark family used for training or evaluation. We ask whether RL post-training can improve abduction as a transferable reasoning capability. We introduce CEDAR-GRPO, a process-aware framework that combines final-answer correctness with abductive rewards for evidence coverage and evidence-to-explanation directionality. Four open-weight LLMs are post-trained on a controlled, domain-neutral mixture of abductive hypothesis-generation and hypothesis-selection tasks. We evaluate them on 11 unseen tasks spanning hypothesis selection, missing-fact generation, defeasible inference, long-context investigation, clinical reasoning, code debugging, and non-abductive controls. CEDAR-GRPO improves every model on every held-out task over both base models and correctness-only GRPO, with average gains of 7.4 and 2.7 points, respectively, and a maximum gain of 30.8 points. Ablations confirm that RL, abductive reward design, and task diversity each contribute to transfer. Process-level metrics further show stronger abductive behavior, including exploration of alternatives, elimination of rivals, backtracking, and uncertainty marking.\footnote{Code and data are available at \url{https://github.com/cedar-grpo/cedar-grpo}.}
\end{abstract}

\section{Introduction}

A clinician interpreting puzzling symptoms, an engineer diagnosing an intermittent failure, and a detective reconstructing a crime from scattered clues all face the same reasoning problem: they must infer a hidden explanation from incomplete evidence. This explanatory step is abductive reasoning, classically understood as moving from observations to a hypothesis that would make them less surprising, and often described as inference to the best explanation \citep{Peirce1931,lipton2004ibe}. For LLMs, abduction is a crucial capability because many real uses of language models---including diagnosis, scientific hypothesis formation, evidence interpretation, debugging, and long-context investigation---depend on identifying plausible latent causes, missing facts, or underlying mechanisms that best account for uncertain observations.

\begin{figure}[!t]
    \centering
    \scriptsize

    \begin{tcolorbox}[
        width=\columnwidth,
        colback=white,
        colframe=black!25,
        boxrule=0.45pt,
        arc=1mm,
        left=4pt,right=4pt,top=4pt,bottom=4pt
    ]
    \raggedright
    \textbf{ART example.} \\
    \textit{Observation 1:} Brad was gazing in the fountain downtown. \\
    \textit{Observation 2:} Then Brad walked away, whistling happily. \\
    \textit{Hypothesis 1:} Brad made a wish on a shooting star. \\
    \textit{Hypothesis 2:} Brad made a wish with a penny. \\
    \textbf{Gold answer:} Hypothesis 2.

    \vspace{0.4em}
    \hrule
    \vspace{0.4em}

    \textbf{Correctness-only GRPO} \\
    Brad was downtown looking into the water of the fountain.
    \textcolor{cedarred}{Fountains have still pools of water that act like mirrors. Since the tall buildings downtown block the view of the sky, he saw a shooting star in the fountain's reflection.}
    He made a wish on that star and walked away whistling because he was happy he caught such a rare sight in the city.
    \textcolor{cedarred}{\textbf{Final Answer: Hypothesis 1.}}

    \vspace{0.4em}
    \hrule
    \vspace{0.4em}

    \textbf{CEDAR-GRPO} \\
    Let us look at the evidence: Brad is ``gazing in the fountain'' and then walks away ``whistling happily.''
    \textcolor{cedargreen}{Public fountains are strongly associated with throwing in a coin to make a wish.
    If we consider Hypothesis 1, Brad would need to be looking upward at the sky.
    But he is specifically gazing in the fountain, so Hypothesis 2 better fits the observations.}
    He likely tossed a penny into the water and walked away happy after making his wish.
    \textcolor{cedargreen}{\textbf{Final Answer: Hypothesis 2.}}
    \end{tcolorbox}

    \caption{CEDAR-GRPO grounds its answer in the observed evidence, while correctness-only GRPO relies on unsupported assumptions.}
    \label{fig:art_qualitative}
\end{figure}

Recent work has introduced valuable benchmarks for abductive reasoning, including ART for commonsense hypothesis selection \citep{bhagavatula2019abductive-commonsense-reasoning}, e-CARE for explainable causal reasoning \citep{du2022ecare}, DDXPlus for differential diagnosis \citep{Tchango2022ddxplus}, ProofWriter and AbductionRules for missing-fact generation \citep{tafjord2020proofwriter,young2022abductionrules}, and knowledge-graph methods for generating hypotheses that explain observed relational structure \citep{bai2023abductive-kg}. However, progress is often evaluated within a single benchmark family, task format, or structured domain, making it difficult to determine whether a method improves abduction as a reusable reasoning capability or merely adapts a model to a narrow formulation. This question is especially important because abductive tasks take two principal forms: hypothesis generation, in which a model proposes an explanation, and hypothesis selection, in which it chooses among candidates \citep{salimi2026wiringwhyunifiedtaxonomy}. A convincing account of abductive improvement should therefore demonstrate transfer across both forms and beyond the specific formats used for training.

\begin{figure*}[t]
    \centering
    \includegraphics[width=0.95\textwidth]{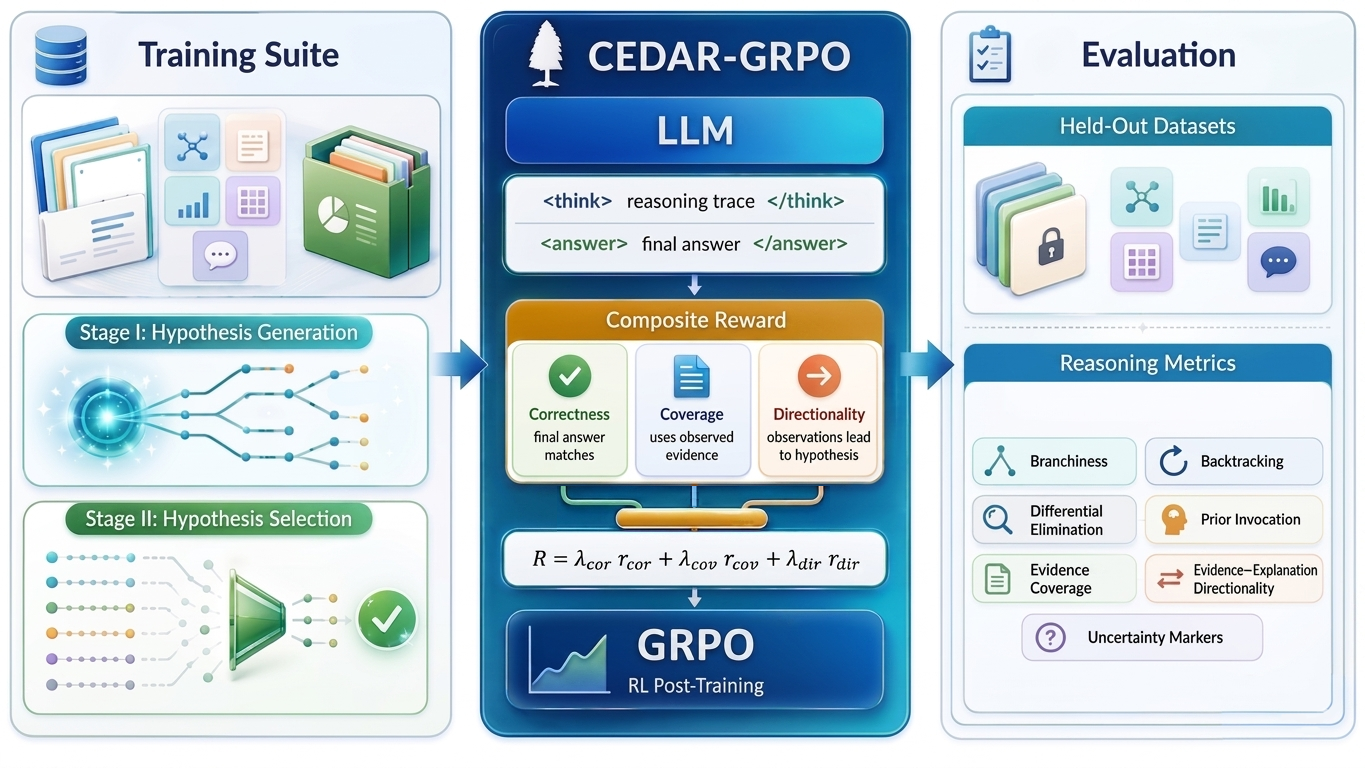}
    \caption{
    Overview of CEDAR-GRPO. The framework trains LLMs on abductive hypothesis generation and selection tasks using structured outputs and a composite reward for correctness, evidence coverage, and evidence--explanation directionality, then evaluates both task accuracy and reasoning behavior on held-out datasets.
    }
    \label{fig:cedar_grpo_overview}
\end{figure*}

We investigate whether reinforcement-learning post-training can produce such transferable improvements. We introduce \textbf{CEDAR-GRPO} (Correctness, Evidence coverage, and Directionality Abductive Rewards), a process-aware GRPO framework that combines deterministic final-answer correctness---using exact-match, label-based, or execution-based verification---with two abductive rewards: coverage of the observed evidence and preservation of the direction from observations to explanation. These rewards are designed to encourage explanations that account for the available evidence while maintaining the defining structure of abductive inference. Figure~\ref{fig:art_qualitative} illustrates the resulting behavior on ART, and Figure~\ref{fig:cedar_grpo_overview} summarizes the training and evaluation pipeline.

We post-train four open-weight backbones---Qwen3-4B, Qwen3-8B, DeepSeek-R1-Distill-Qwen-7B, and Llama-3.1-8B-Instruct---on a controlled, domain-neutral, knowledge-light mixture spanning both hypothesis generation and hypothesis selection. The mixture includes formal and commonsense hypothesis selection, missing-fact generation, and rule-learning tasks. This design makes transfer easier to interpret: improvements outside these training formats are less plausibly attributable to narrow domain adaptation.

We evaluate on 11 unseen tasks covering hypothesis selection, missing-fact generation, defeasible inference, long-context investigation, clinical reasoning, code debugging, forward causal reasoning, and general multistep reasoning. The suite contains direct abductive targets, abduction-adjacent transfer tasks, and non-abductive controls. We also analyze reasoning traces using seven process-level metrics---branchiness, backtracking, differential elimination, prior invocation, evidence coverage, evidence--explanation directionality, and uncertainty marking---to test whether accuracy gains coincide with changes in abductive behavior.

Across all four backbones, CEDAR-GRPO improves performance on every held-out task relative to both the base models and the correctness-only GRPO baseline, Cor-GRPO. Averaged across settings, it gains 7.4 points over the base models and 2.7 points over Cor-GRPO, with a maximum gain of 30.8 points over the base model on MuSR-Murder for DeepSeek-R1-Distill-Qwen-7B. As expected, evidence coverage and directionality, both directly optimized during training, improve substantially. Beyond these reward metrics, CEDAR-GRPO also improves all five held-out process metrics, showing greater exploration of alternatives, elimination of competing hypotheses, backtracking, prior invocation, and uncertainty marking. Ablations indicate that RL, both process-reward components, and the combination of generation- and selection-oriented training data each contribute to transfer.

Taken together, these results show that abductive reasoning can be strengthened as a \textbf{transferable capability}, rather than through benchmark-specific optimization alone. Our main contributions are: (1) a \textbf{controlled framework for studying cross-task abductive transfer}; (2) \textbf{CEDAR-GRPO}, which augments correctness-based RL with rewards for evidence coverage and evidence--explanation directionality; and (3) \textbf{systematic evidence of transfer} across four backbones and 11 held-out tasks, supported by task-level evaluation, held-out process metrics, and targeted ablations.

\section{Related Work}
\label{sec:related_work}

\subsection{Abductive Tasks and Benchmarks}

Abductive reasoning in NLP is commonly studied through tasks that ask models to infer plausible explanations, missing facts, or latent causes from incomplete observations. Existing benchmarks instantiate this ability in several forms, including commonsense hypothesis selection in ART and UNcommonsense, explainable causal reasoning in e-CARE, clinical differential diagnosis in DDXPlus, abductive missing-fact generation in ProofWriter and AbductionRules, and structured hypothesis generation over knowledge graphs \citep{bhagavatula2019abductive-commonsense-reasoning,zhao2023uncommonsense,du2022ecare,Tchango2022ddxplus,tafjord2020proofwriter,young2022abductionrules,bai2023abductive-kg}. These resources have been essential for measuring abductive behavior, but they often make improvements hard to disentangle from particular domains, output formats, or verification regimes. Our work therefore treats abduction as a broader capability whose improvement should transfer across both hypothesis-generation and hypothesis-selection settings, as well as beyond the tasks used for training.

\subsection{Methodologies for Abductive Reasoning}

Prior methods for improving abductive reasoning have largely relied on supervised adaptation or inference-time scaffolding. Supervised, ranking-based, and preference-based objectives train models from labeled abductive data, either by imitating reference explanations or by learning to prefer stronger hypotheses \citep{zhu2020l2r2,he2025gear}. In contrast, recent LLM pipelines use prompting, retrieval, multi-agent decomposition, or symbolic checking to separate observation interpretation, hypothesis generation, and hypothesis evaluation \citep{liu2024incomplete-loop-instruction-inference,lin2025abductiveinferenceretrievalaugmentedlanguage,he2023lego,li2025hypothesis,hong2024argmed}. Recently, LogiDynamics \citep{zheng2025logidynamics} demonstrated that embedding abductive inference within a broader logical-reasoning pipeline alongside iterative refinement systematically enhances performance. While effective for task-specific adaptation or inference control, these approaches leave open whether post-training can internalize broadly generalizable abductive behaviors.


\subsection{RL Post-Training and Process-Aware Rewards}

Reinforcement learning offers a natural way to optimize abductive reasoning against task-level criteria, but its use in abduction remains narrow. RLF-KG uses PPO-style feedback from knowledge graphs to generate logical hypotheses that explain observed facts, while CtrlHGen applies GRPO-based reinforcement tuning with rewards for semantic alignment and condition satisfaction, and DARK applies a coupled-GRPO variant to a masked diffusion model for unifying deductive and abductive reasoning over knowledge graphs \citep{bai2023abductive-kg,gao2025ctrlhgen,gao2026unifyingdeductiveabductivereasoning}. More broadly, GRPO was introduced for mathematical reasoning as a memory-efficient alternative to PPO \citep{shao2024deepseekmath}, and recent RL post-training work shows that verifiable rewards can elicit behaviors such as verification, reflection, and strategy adaptation \citep{guo2025deepseekr1}. However, these successes are concentrated in math, code, or structured reasoning. CEDAR-GRPO instead studies broad abductive transfer: it combines verifiable correctness with process-aware rewards for evidence coverage and evidence--explanation directionality, drawing inspiration from process-supervision work while avoiding reliance on human-labeled reasoning traces \citep{lightman2024lets,uesato2022solving}.

\section{Data Collection}
\label{sec:data_collection}

\subsection{Dataset selection}
We construct both the training pool and evaluation suite to probe abductive reasoning in a controlled yet heterogeneous manner. Following the two-stage view of abduction adopted in recent survey work, we treat abductive reasoning as comprising \emph{Stage~I} hypothesis generation and \emph{Stage~II} hypothesis selection \citep{salimi2026wiringwhyunifiedtaxonomy}. This distinction is crucial in our setting: a model can improve by proposing better explanatory hypotheses, by evaluating candidate hypotheses more reliably, or by doing both. To avoid conflating these possibilities, we do not build the training pool around a single benchmark family.

Our data collection strategy is intentionally asymmetric. The training pool is limited to broadly non-specialized resources and spans both stages of the abductive pipeline, so that gains are less likely to reflect narrow domain adaptation or isolated progress on either generation or selection. The evaluation suite is broader: alongside classic abductive benchmarks, it includes abstract logical data, neighboring tasks that admit an abductive interpretation, and explicit non-abductive controls. This design lets us ask a sharper question: does GRPO learn reusable abductive behavior, or merely adapt to the surface form of a few benchmarks? Throughout, we maintain a strict separation between learning and measurement: evaluation datasets are never used for model optimization or model selection. Additional dataset-specific notes are deferred to Appendix~\ref{app:data_collection}.

\subsection{Training and validation data}

Our training pool, detailed in Appendix~\ref{app:data_collection} (Table~\ref{tab:train_data}), is composed of datasets targeting both Stage~I and Stage~II of abduction. The Stage~II group requires the model to assess the plausibility of an explanation, cause, or evidential relation, or to select among competing hypotheses. UniADILR-HGc and Balanced COPA provide compact classic abductive selection signals. CauseLogics asks whether a candidate premise makes an inference sensible, capturing a core part of hypothesis assessment: checking whether a hypothesis can logically and soundly explain the observations. CLIMATE-FEVER similarly provides a Stage~II-adjacent signal, requiring the model to choose among supports, refutes, not enough info, and disputed for a claim--evidence relation. The Stage~I group instead requires constructing missing hypotheses rather than choosing from fixed alternatives. AbductionRules asks the model to generate a missing explanatory fact from a rule context, while List Function and Crypto require inferring a latent pattern from input-output examples and generating code that implements it; the code can therefore be interpreted as an explicit candidate hypothesis about the governing rule, making these tasks useful proxies for hypothesis generation.

This composition is useful for GRPO in two ways. Selection-style tasks provide clean signals for plausibility judgments, while generation-style tasks require the model to construct missing hypotheses rather than only evaluate a closed set of candidates. Mixing both makes it less likely that post-training improves only a narrow answer-selection heuristic. Just as importantly, the training sources remain non-specialized, which makes the subsequent evaluation cleaner: if performance improves on downstream tasks from other settings, the gain is less plausibly explained by domain memorization.

\subsection{Evaluation data}

Our evaluation suite, detailed in Appendix~\ref{app:data_collection} (Table~\ref{tab:eval_data}), is deliberately broader than the training pool. It includes classic abductive benchmarks---ART ($\alpha$NLI) for Stage~II hypothesis selection and NeuLR's abductive split for Stage~I missing-fact generation---as well as Defeasible NLI, which tests the Stage~II-adjacent ability to judge whether new evidence strengthens or weakens a hypothesis.

We also evaluate transfer to neighboring domains that admit abductive interpretations: GoEmotions as latent affect inference from utterances, MuSR: Murder as culprit inference, MedQA as clinical diagnosis, and ML-debugging as software diagnosis and repair. Finally, we include non-abductive controls: Balanced COPA in the effect direction for forward causal reasoning, MuSR: Object and MuSR: Team for long-context multistep reasoning, and StrategyQA as a broad general-reasoning control. Together, these groups help distinguish direct abductive transfer from broader changes in reasoning behavior, while preserving a strict separation between training sources and held-out evaluation.

\section{Methodology}

\subsection{Problem Formulation}

We operationalize abductive reasoning as selecting or generating the explanation that best accounts for incomplete observations. Across the training mixture, the target explanation may be a candidate cause, a missing fact, or an inferred transformation rule. Our experiments use four open-weight backbones: Qwen3-4B, Qwen3-8B, DeepSeek-R1-Distill-Qwen-7B, and Llama-3.1-8B-Instruct. This model suite spans 4B--8B parameters and includes both general instruction-tuned and reasoning-oriented backbones, allowing us to test whether our method improves abductive reasoning across model sizes and families.

\subsection{Structured CoT Prompting}

Models generate structured outputs of the form
$\langle\texttt{think}\rangle\,\beta\,\langle\texttt{/think}\rangle
\langle\texttt{answer}\rangle\,\alpha\,\langle\texttt{/answer}\rangle$,
where $\beta$ is the reasoning trace and $\alpha$ is the final answer.
Task correctness is computed from $\alpha$, whereas the process rewards are
computed from the user prompt and $\beta$. Full generator prompts are
provided in Appendix~\ref{app:prompts}.

\subsection{Composite Reward Design}
\label{sec:composite_reward}

For a user prompt $x$, generated completion $y$, ground truth $g$, and dataset $d$, let $\beta(y)=\operatorname{ExtractThink}(y)$ and $\alpha(y)=\operatorname{ExtractAnswer}(y)$. CEDAR-GRPO assigns
\begin{equation}
R(x,y,g,d)=\frac{r_{\mathrm{cor}}+r_{\mathrm{cov}}+r_{\mathrm{dir}}}{3}.
\label{eq:cedar_reward}
\end{equation}
The correctness reward is
\begin{equation}
r_{\mathrm{cor}}=V_d(\alpha(y),g),
\qquad
V_d(\alpha(y),g)\in\{0,1\},
\end{equation}
where $V_d$ is the dataset-specific exact-match, set-match, or
execution-based verifier. For coverage, the judge returns $m$ observation details from $x$ and marks
each as addressed ($z_j=1$) or unaddressed ($z_j=0$):
\begin{equation}
r_{\mathrm{cov}}=\frac{1}{m}\sum_{j=1}^{m} z_j.
\end{equation}
If no valid detail list is returned, $r_{\mathrm{cov}}$ is set to zero.

The directionality reward is $0$ for traces that assume an explanation and
reason back toward the evidence, $0.5$ for mixed or ambiguous
directionality, and $1$ for traces that reason from evidence toward an
explanation. Both process rewards are scored by \texttt{gpt-oss-120b} at temperature $0.0$ from the user prompt $x$ and reasoning span $\beta(y)$; correctness is computed only from $\alpha(y)$. Exact verifiers, failure handling, dataset notes, and
training-time judge prompts are given in
Appendix~\ref{sec:appendix_rewardfn}.

Figure~\ref{fig:reward_progression} illustrates how three traces with the
same correct answer receive different rewards because they differ in
evidence coverage and reasoning directionality. We selected these process
terms because they apply across the full training mixture and are less
susceptible than alternative trace metrics to simple reward-hacking
strategies such as verbosity and hedging. The screening analysis is
reported in Appendix~\ref{app:reward_selection}.

\subsection{GRPO Optimization and Compute}

For each prompt, we sample $G=4$ completions and optimize Equation~\ref{eq:cedar_reward} with Group Relative Policy Optimization (GRPO), which forms group-relative advantages without a separate value network. All backbones use NF4 quantization and LoRA fine-tuning. All local training and evaluation were conducted on a single NVIDIA GeForce RTX~5090 GPU with 32~GB GDDR7; the reward judge was accessed remotely. Full hyperparameters are reported in Appendix~\ref{sec:appendix_implementation}.

\begin{figure*}[!t]
    \centering
    \begingroup
    \definecolor{rewardcor}{RGB}{112,74,168}
    \definecolor{rewardcov}{RGB}{34,104,165}
    \definecolor{rewarddir}{RGB}{20,120,70}
    \definecolor{rewardbad}{RGB}{165,45,45}

    \newcommand{\cormark}[1]{\textcolor{rewardcor}{\textbf{#1}}}
    \newcommand{\covmark}[1]{\textcolor{rewardcov}{\textbf{#1}}}
    \newcommand{\dirmark}[1]{\textcolor{rewarddir}{\textbf{#1}}}
    \newcommand{\revmark}[1]{\textcolor{rewardbad}{\textbf{#1}}}
    \newcommand{\rewardchip}[3]{%
        \begingroup\setlength{\fboxsep}{2pt}%
        \colorbox{#1!10}{\textcolor{#1}{\textbf{#2}~#3}}%
        \endgroup}
    \newcommand{\detailon}[1]{%
        \begingroup\setlength{\fboxsep}{1.5pt}%
        \colorbox{rewardcov!12}{\textcolor{rewardcov}{\textbf{#1}}}%
        \endgroup}
    \newcommand{\detailoff}[1]{%
        \begingroup\setlength{\fboxsep}{1.5pt}%
        \colorbox{black!5}{\textcolor{black!45}{\textbf{#1}}}%
        \endgroup}

    \scriptsize
    \begin{tcolorbox}[
        width=\textwidth,
        colback=white,
        colframe=black!25,
        boxrule=0.5pt,
        arc=1mm,
        left=4pt,right=4pt,top=4pt,bottom=4pt
    ]
    \raggedright
    \textbf{BalancedCOPA example (cause selection).}
    \textit{Premise:} ``The investigator found inconsistencies in the witness's story.''
    \quad
    \textit{Option 1:} The witness fabricated parts of the account.
    \quad
    \textit{Option 2:} The investigator was reassigned.
    \quad
    \textbf{Gold: 1.}

    \vspace{0.35em}
    \textbf{Coverage judge decomposes the premise into:}
    \detailon{A} inconsistencies exist;
    \detailon{B} they are in the witness's story;
    \detailon{C} the investigator detects them.
    \hfill
    $R=(r_{\mathrm{cor}}+r_{\mathrm{cov}}+r_{\mathrm{dir}})/3$.

    \vspace{0.35em}
    \hrule
    \vspace{0.45em}

    \begin{center}
    \textbf{Training progression}\hspace{0.5em}
    $\longrightarrow$\hspace{0.5em}
    \textit{correct answer} $\rightarrow$ \textit{grounded abductive explanation}
    \end{center}
    \vspace{-0.35em}

    \noindent
    \begin{minipage}[t]{0.322\linewidth}
    \vspace{0pt}
    \begin{tcolorbox}[
        enhanced,
        equal height group=rewardepochs,
        valign=top,
        colback=rewardbad!2,
        colframe=rewardbad!35,
        boxrule=0.45pt,
        arc=1mm,
        title={\textbf{Epoch 1} \hfill Step 48},
        colbacktitle=rewardbad!9,
        coltitle=black,
        fonttitle=\scriptsize,
        left=3pt,right=3pt,top=3pt,bottom=3pt
    ]
    \raggedright
    \textbf{Generator trace (condensed)}\\[-0.1em]
    \texttt{<think>} Option 1 says the witness lied; Option 2 is about a reassignment.
    \revmark{If someone lies, their story has problems.}
    So it is Option 1. \texttt{</think>}\\
    \cormark{\texttt{<answer>1</answer>}}

    \vspace{0.45em}
    \textbf{Judge view.}
    ``Problems'' is a generic consequence, not an explicit explanation of any premise detail. The trace also starts from an assumed lie and predicts the observation.

    \vspace{0.45em}
    \textbf{Explicitly linked details:}
    \detailoff{A}\,\detailoff{B}\,\detailoff{C}

    \vfill
    \hrule
    \vspace{0.45em}
    \centering
    \rewardchip{rewardcor}{$r_{\mathrm{cor}}$}{1.00}\quad
    \rewardchip{rewardcov}{$r_{\mathrm{cov}}$}{0/3}
    \\\vspace{0.3em}
    \rewardchip{rewarddir}{$r_{\mathrm{dir}}$}{0.00}\quad
    \rewardchip{black}{$R$}{\textbf{0.33}}
    \end{tcolorbox}
    \end{minipage}\hfill
    \begin{minipage}[t]{0.322\linewidth}
    \vspace{0pt}
    \begin{tcolorbox}[
        enhanced,
        equal height group=rewardepochs,
        valign=top,
        colback=rewardcov!2,
        colframe=rewardcov!35,
        boxrule=0.45pt,
        arc=1mm,
        title={\textbf{Epoch 2} \hfill Step 534},
        colbacktitle=rewardcov!8,
        coltitle=black,
        fonttitle=\scriptsize,
        left=3pt,right=3pt,top=3pt,bottom=3pt
    ]
    \raggedright
    \textbf{Generator trace (condensed)}\\[-0.1em]
    \texttt{<think>} The observation is that the investigator found inconsistencies in the witness's story.
    \dirmark{I need to find what most plausibly caused this.}

    Option 1: Fabrication means inventing or altering details. Combining fabricated and truthful sections creates
    \covmark{internal contradictions (A)} in the
    \covmark{witness's account (B)}---exactly what ``inconsistencies'' describes.

    Option 2: Reassignment changes workload, not story content, and gives no mechanism for inconsistencies. Ruled out. \texttt{</think>}\\
    \cormark{\texttt{<answer>1</answer>}}

    \vspace{0.45em}
    \textbf{Judge view.}
    The trace begins from the observation and explicitly explains A and B. It mentions detection, but does not yet explain why the inconsistency is discoverable by the investigator (C).

    \vspace{0.45em}
    \textbf{Explicitly linked details:}
    \detailon{A}\,\detailon{B}\,\detailoff{C}

    \vfill
    \hrule
    \vspace{0.45em}
    \centering
    \rewardchip{rewardcor}{$r_{\mathrm{cor}}$}{1.00}\quad
    \rewardchip{rewardcov}{$r_{\mathrm{cov}}$}{2/3}
    \\\vspace{0.3em}
    \rewardchip{rewarddir}{$r_{\mathrm{dir}}$}{1.00}\quad
    \rewardchip{black}{$R$}{\textbf{0.89}}
    \end{tcolorbox}
    \end{minipage}\hfill
    \begin{minipage}[t]{0.322\linewidth}
    \vspace{0pt}
    \begin{tcolorbox}[
        enhanced,
        equal height group=rewardepochs,
        valign=top,
        colback=rewarddir!2,
        colframe=rewarddir!35,
        boxrule=0.45pt,
        arc=1mm,
        title={\textbf{Epoch 3} \hfill Step 1,018},
        colbacktitle=rewarddir!8,
        coltitle=black,
        fonttitle=\scriptsize,
        left=3pt,right=3pt,top=3pt,bottom=3pt
    ]
    \raggedright
    \textbf{Generator trace (condensed)}\\[-0.1em]
    \texttt{<think>} Observation: the investigator found inconsistencies in the witness's story.
    \dirmark{I reason abductively---from this observation toward its most plausible cause.}

    Option 1: Fabrication
    \covmark{creates traceable contradictions (A)},
    \covmark{localizes the error in the witness's narrative (B)}, and
    \covmark{unravels under systematic questioning, explaining why the investigator detects it (C)}.
    All three facts are parsimoniously explained.

    Option 2: Reassignment affects caseload, not story content; it has no causal path to the contradictions and would more naturally explain missed, not found, inconsistencies. Decisively ruled out.

    Conclusion: Option 1 uniquely explains the evidence. \texttt{</think>}\\
    \cormark{\texttt{<answer>1</answer>}}

    \vspace{0.45em}
    \textbf{Explicitly linked details:}
    \detailon{A}\,\detailon{B}\,\detailon{C}

    \vfill
    \hrule
    \vspace{0.45em}
    \centering
    \rewardchip{rewardcor}{$r_{\mathrm{cor}}$}{1.00}\quad
    \rewardchip{rewardcov}{$r_{\mathrm{cov}}$}{3/3}
    \\\vspace{0.3em}
    \rewardchip{rewarddir}{$r_{\mathrm{dir}}$}{1.00}\quad
    \rewardchip{black}{$R$}{\textbf{1.00}}
    \end{tcolorbox}
    \end{minipage}

    \vspace{0.4em}
    \noindent
    \rewardchip{rewardcor}{Purple}{correct final answer}\quad
    \rewardchip{rewardcov}{Blue}{explicit evidence--hypothesis link}\quad
    \rewardchip{rewarddir}{Green}{evidence $\rightarrow$ explanation}\quad
    \rewardchip{rewardbad}{Red}{explanation $\rightarrow$ evidence}
    \end{tcolorbox}
    \endgroup

    \caption{How CEDAR-GRPO's composite reward shapes a BalancedCOPA reasoning trace over training. All three snapshots select the correct answer, so $r_{\mathrm{cor}}$ is already maximal. At Step~48, the trace reasons from an assumed explanation back to a generic consequence and accounts for none of the three observation details returned by the coverage judge. By Step~534, it starts from the observation and links two details to the hypothesis. By Step~1,018, it explains all three details while preserving the evidence-to-explanation direction, yielding the maximum reward.}
    \label{fig:reward_progression}
\end{figure*}

\section{Experiments}
\label{sec:experiments}

This section evaluates whether post-training with our composite reward improves general abductive reasoning, rather than simply increasing performance on the training tasks. We organize the evidence into three parts: (i) held-out task performance on the evaluation suite from Section~\ref{sec:data_collection}; (ii) process-level measurements of the reasoning traces; and (iii) ablations that isolate the role of RL, reward design, the two-stage construction of the training pool, and a matched generic-reasoning post-training control. We use accuracy for closed-form selection tasks, exact or verifier-based correctness for formal missing-fact and rule-learning tasks, and pass/fix success for ML-debugging. All reported scores are computed from the final answer field only; process metrics are computed separately from the reasoning trace.

\subsection{Held-Out Task Performance}
\label{sec:heldout_results}

Table~\ref{tab:main_results} presents the main task-level comparison. We compare the original base model, the correctness-only GRPO checkpoint (denoted Cor-GRPO), and our main composite-reward checkpoint, CEDAR-GRPO.

\begin{table*}[t]
\centering
\small
\setlength{\tabcolsep}{3pt}
\renewcommand{\arraystretch}{1.08}
\resizebox{\textwidth}{!}{%
\begin{tabular}{llccccccccccc}
\toprule
Model & Method & ART & B-COPA & DefNLI & GoEmo. & MuSR-M & MuSR-O & MuSR-T & NeuLR & StratQA & MedQA & ML-Debug \\
\midrule
\multirow{3}{*}{Qwen3-4B}
& Base & $65.25\%$ & $84.40\%$ & $80.75\%$ & $27.75\%$ & $20.40\%$ & $19.20\%$ & $51.60\%$ & $35.00\%$ & $38.25\%$ & $35.50\%$ & $25.25\%$ \\
& Cor-GRPO & $71.75\%$ & $86.80\%$ & $86.75\%$ & $32.25\%$ & $41.60\%$ & $32.40\%$ & $48.80\%$ & $38.75\%$ & $43.25\%$ & $37.50\%$ & $28.50\%$ \\
& CEDAR-GRPO & $\mathbf{72.25\%}$ & $\mathbf{88.80\%}$ & $\mathbf{88.50\%}$ & $\mathbf{34.25\%}$ & $\mathbf{48.40\%}$ & $\mathbf{33.20\%}$ & $\mathbf{55.60\%}$ & $\mathbf{40.50\%}$ & $\mathbf{44.75\%}$ & $\mathbf{37.75\%}$ & $\mathbf{29.00\%}$ \\
\midrule
\multirow{3}{*}{Qwen3-8B}
& Base & $72.25\%$ & $84.40\%$ & $82.50\%$ & $39.50\%$ & $23.20\%$ & $26.40\%$ & $55.20\%$ & $36.25\%$ & $40.00\%$ & $42.25\%$ & $28.75\%$ \\
& Cor-GRPO & $74.00\%$ & $86.00\%$ & $87.75\%$ & $45.50\%$ & $43.60\%$ & $34.00\%$ & $58.80\%$ & $39.50\%$ & $44.50\%$ & $46.25\%$ & $29.75\%$ \\
& CEDAR-GRPO & $\mathbf{75.50\%}$ & $\mathbf{89.20\%}$ & $\mathbf{90.50\%}$ & $\mathbf{47.00\%}$ & $\mathbf{46.80\%}$ & $\mathbf{36.80\%}$ & $\mathbf{60.80\%}$ & $\mathbf{43.75\%}$ & $\mathbf{50.50\%}$ & $\mathbf{49.25\%}$ & $\mathbf{31.75\%}$ \\
\midrule
\multirow{3}{2.7cm}{\raggedright DeepSeek-R1-Distill-Qwen-7B}
& Base & $70.25\%$ & $85.20\%$ & $81.50\%$ & $30.00\%$ & $26.40\%$ & $38.40\%$ & $49.60\%$ & $31.50\%$ & $42.50\%$ & $35.75\%$ & $23.75\%$ \\
& Cor-GRPO & $73.50\%$ & $88.00\%$ & $82.25\%$ & $31.50\%$ & $56.40\%$ & $40.40\%$ & $50.00\%$ & $33.00\%$ & $42.25\%$ & $36.25\%$ & $22.75\%$ \\
& CEDAR-GRPO & $\mathbf{78.50\%}$ & $\mathbf{89.60\%}$ & $\mathbf{87.75\%}$ & $\mathbf{35.50\%}$ & $\mathbf{57.20\%}$ & $\mathbf{49.60\%}$ & $\mathbf{51.60\%}$ & $\mathbf{35.50\%}$ & $\mathbf{45.75\%}$ & $\mathbf{39.50\%}$ & $\mathbf{24.75\%}$ \\
\midrule
\multirow{3}{2.7cm}{\raggedright Llama-3.1-8B-Instruct}
& Base & $73.50\%$ & $86.40\%$ & $86.00\%$ & $33.75\%$ & $24.80\%$ & $35.20\%$ & $48.80\%$ & $29.00\%$ & $42.50\%$ & $33.50\%$ & $19.00\%$ \\
& Cor-GRPO & $78.75\%$ & $89.20\%$ & $87.75\%$ & $35.50\%$ & $52.00\%$ & $36.00\%$ & $49.20\%$ & $30.25\%$ & $42.25\%$ & $34.25\%$ & $19.75\%$ \\
& CEDAR-GRPO & $\mathbf{80.00\%}$ & $\mathbf{90.40\%}$ & $\mathbf{91.50\%}$ & $\mathbf{39.25\%}$ & $\mathbf{53.20\%}$ & $\mathbf{36.80\%}$ & $\mathbf{50.80\%}$ & $\mathbf{32.25\%}$ & $\mathbf{45.50\%}$ & $\mathbf{35.75\%}$ & $\mathbf{21.25\%}$ \\
\bottomrule
\end{tabular}%
}
\caption{Held-out task performance across the evaluation suite. Scores represent task-native accuracy or exact-success metrics. CEDAR-GRPO is our composite-reward method; Cor-GRPO is the correctness-only baseline. The best score per backbone is bolded. Full dataset descriptions and abbreviations are provided in Appendix~\ref{app:data_collection}.}
\label{tab:main_results}
\end{table*}

This held-out suite tests transfer beyond the training formats, spanning direct abductive tasks, abduction-adjacent settings such as clinical diagnosis and ML debugging, and non-abductive controls. Comparing against both the base model and Cor-GRPO separates gains from RL itself from gains due to the composite reward. The results show that CEDAR-GRPO consistently improves over Cor-GRPO, suggesting that the process-level signals are not merely cosmetic. Instead, the coverage and directionality rewards are aligned with the underlying reasoning objective and lead to better final-answer accuracy.

\subsection{Reasoning Trace Analysis}
\label{sec:trace_analysis}

\begin{table}[t]
  \centering
  \small
  \setlength{\tabcolsep}{4pt}
  \begin{tabularx}{\linewidth}{@{}p{2.3cm} X@{}}
    \toprule
    \textbf{Metric} & \textbf{Description} \\
    \midrule
    \textbf{Branchiness} & \raggedright\arraybackslash Exploring multiple distinct candidate explanations for the same observation. \\
    \addlinespace
    \textbf{Backtracking} & \raggedright\arraybackslash Explicitly identifying an error or flaw in the reasoning and changing direction. \\
    \addlinespace
    \textbf{Differential \newline Elimination} & \raggedright\arraybackslash The active refutation of alternative hypotheses given the specific context. \\
    \addlinespace
    \textbf{Prior Invocation} & \raggedright\arraybackslash Incorporating typicality or prior probability alongside case-specific evidence. \\
    \addlinespace
    \textbf{Evidence \newline Coverage} & \raggedright\arraybackslash The fraction of specific observation details explicitly accounted for by the chosen hypothesis. \\
    \addlinespace
    \textbf{Evidence--Expl. \newline Directionality} & \raggedright\arraybackslash Demonstrating awareness that reasoning must move from given evidence toward an explanatory conclusion. \\
    \addlinespace
    \textbf{Uncertainty \newline Markers} & \raggedright\arraybackslash The density of probabilistic language and epistemic hedging within the trace. \\
    \bottomrule
  \end{tabularx}
  \caption{Overview of the process-level metrics used to evaluate intermediate reasoning behaviors.}
  \label{tab:metrics-taxonomy}
\end{table}

While our training improves accuracy on held-out benchmarks, correct final answers alone do not guarantee that the underlying reasoning process has improved. Because final-answer accuracy is a coarse outcome measure, we complement it with process-level evaluation to examine whether the model's reasoning chains show the abductive behaviors targeted by our training objective. To do this, we introduce a novel suite of process-level metrics, summarized in Table~\ref{tab:metrics-taxonomy} and discussed in more detail in Appendix~\ref{app:process_level_metrics}, that directly evaluate intermediate reasoning chains. These metrics assess whether the model generates plausible hypotheses, connects them to the available evidence, and uses them to support the final answer. This gives more concrete evidence that the accuracy gains are accompanied by meaningful improvements in the model's reasoning behavior. The reasoning trace analysis in this work is conducted using DeepSeek-R1-Distill-Qwen-7B.



\begin{table*}[tbp]
  \centering
  \small
  \setlength{\tabcolsep}{3pt} 
  \renewcommand{\arraystretch}{1.2}
  \begin{tabular}{l !{\hskip 4pt} c !{\hskip 4pt} c !{\hskip 4pt} c !{\hskip 4pt} c !{\hskip 4pt} c !{\hskip 12pt} c !{\hskip 4pt} c}
    \toprule
    & \multicolumn{5}{c!{\hskip 12pt}}{\textbf{Held-out Metrics}} & \multicolumn{2}{c}{\textbf{Reward Metrics}} \\
    \cmidrule(r{12pt}){2-6} \cmidrule{7-8}
    \textbf{Method} & \textbf{Backtracking} & \textbf{Branchiness} & \textbf{Diff. Elim.} & \textbf{Prior} & \textbf{Uncertainty} & \textbf{Coverage} & \textbf{Direction.} \\
    \midrule
    Baseline & 0.69 & 1.22 & 0.79 & 0.59 & 0.87 & 33.1\% & 0.21 \\
    \addlinespace
    Cor-GRPO & \cellcolor[HTML]{A2DCBA}0.93 & \cellcolor[HTML]{F5DFDD}1.16 & \cellcolor[HTML]{B3E3C7}0.97 & \cellcolor[HTML]{F0D1CD}0.53 & \cellcolor[HTML]{D9F1E3}0.92 & \cellcolor[HTML]{BCE6CD}39.1\% & \cellcolor[HTML]{E8B8B3}0.16 \\
    \addlinespace
    \textbf{CEDAR-GRPO} & \cellcolor[HTML]{86D2A6}\textbf{1.09} & \cellcolor[HTML]{AFE1C4}\textbf{1.53} & \cellcolor[HTML]{81D0A2}\textbf{1.29} & \cellcolor[HTML]{B5E3C8}\textbf{0.72} & \cellcolor[HTML]{87D2A7}\textbf{1.37} & \cellcolor[HTML]{84D1A5}\textbf{52.9\%} & \cellcolor[HTML]{27AE60}\textbf{0.60} \\
    \bottomrule
  \end{tabular}
  \caption{Mean process-level metric scores averaged across ten held-out evaluation datasets for DeepSeek-R1-Distill-Qwen-7B. Evidence Coverage and Directionality were explicitly included as rewards in the CEDAR-GRPO training objective; the other five metrics are fully held-out metrics used to assess reasoning generalization. Shading indicates a relative increase ($\color[HTML]{27ae60}\textbf{green}$) or decrease ($\color[HTML]{c0392b}\textbf{red}$) against the Baseline. Full per-dataset breakdowns are provided in Appendix~\ref{app:process_level_metrics}, Table~\ref{tab:model_results}.}
  \label{tab:mean_heatmap_table}
\end{table*}

As shown in Table~\ref{tab:mean_heatmap_table}, optimizing with the composite reward (CEDAR-GRPO) understandably leads to substantial increases in the two metrics it directly targets: Evidence Coverage and Evidence--Explanation Directionality. Evidence Coverage increases substantially from the baseline of 33.1\% to 52.9\%, and Directionality rises from 0.21 to 0.60. Interestingly, correctness-only optimization (Cor-GRPO) struggles to maintain this structural rigor, with Directionality dropping slightly to 0.16. This demonstrates that without explicit grounding, the model may arrive at correct answers without a logically sound evidence-to-conclusion flow.

While the gains in Coverage and Directionality confirm that the optimization successfully induced the specific rewarded behaviors, our central finding is the significant positive transfer to the fully held-out metrics. Crucially, we observe consistent improvements in exploratory behaviors that were not explicitly rewarded during training. Backtracking and Differential Elimination show consistent increases relative to the baseline under both training objectives, indicating that some degree of error correction and active refutation naturally emerges from RL training. The composite reward, however, amplifies these traits considerably, pushing Differential Elimination from 0.79 to 1.29 and Backtracking from 0.69 to 1.09.

A key divergence between the two training regimes appears in how the model generates and explores candidate explanations. Under correctness-only optimization, the reasoning traces become narrower: Branchiness decreases from 1.22 to 1.16, and Prior Invocation drops from 0.59 to 0.53. In contrast, the composite reward encourages the model to actively hypothesize and explore multiple distinct paths, raising Branchiness to 1.53 and Prior Invocation to 0.72. This suggests that CEDAR-GRPO induces a broader shift toward exploratory, transferable reasoning rather than narrow optimization for final answers.

Finally, while the baseline model already exhibits some use of uncertainty markers (0.87), composite training leads to a considerable increase, reaching 1.37. This trend suggests that rather than simply generating verbose filler, the model learns to qualify its steps and explicitly acknowledge epistemic uncertainty when evaluating alternative explanations.

Together, these metrics show that the composite reward improves the model's abductive reasoning behavior, encouraging it to actively hypothesize and evaluate alternatives rather than passively converging on a correct final answer.

\subsection{Ablation Studies}
\label{sec:ablations}

In this section we test whether the observed gains are caused by abductive RL itself, by exposure to the same data under a supervised objective, by the composition of the reward, by the two-stage structure of the training pool, or by generic reasoning post-training. To keep the ablation grid tractable, all ablations are run only on Qwen3-4B and DeepSeek-R1-Distill-Qwen-7B. Table~\ref{tab:unified_ablations} summarizes all ablation variants; the following subsections refer to the corresponding row groups.

\begin{table*}[t]
\centering
\small
\setlength{\tabcolsep}{2.5pt}
\renewcommand{\arraystretch}{1.08}
\resizebox{\textwidth}{!}{%
\begin{tabular}{lllcccccccccccc}
\toprule
Model & Ablation & Method & ART & B-COPA & DefNLI & GoEmo. & MuSR-M & MuSR-O & MuSR-T & NeuLR & StratQA & MedQA & ML-Debug & Avg. $\Delta$ vs Base \\
\midrule
\multirow{9}{*}{Qwen3-4B} 
& Reference & Base
& $65.25\%$ & $84.40\%$ & $80.75\%$ & $27.75\%$ & $20.40\%$ & $19.20\%$ & $51.60\%$ & $35.00\%$ & $38.25\%$ & $35.50\%$ & $25.25\%$ & $0.00$ \\
& Reference & Cor-GRPO
& $71.75\%$ & $86.80\%$ & $86.75\%$ & $32.25\%$ & $41.60\%$ & $32.40\%$ & $48.80\%$ & $38.75\%$ & $43.25\%$ & $37.50\%$ & $28.50\%$ & $+5.91$ \\
& Main & \textbf{CEDAR-GRPO}
& $\mathbf{72.25\%}$ & $\mathbf{88.80\%}$ & $\mathbf{88.50\%}$ & $34.25\%$ & $\mathbf{48.40\%}$ & $\mathbf{33.20\%}$ & $\mathbf{55.60\%}$ & $\mathbf{40.50\%}$ & $44.75\%$ & $\mathbf{37.75\%}$ & $\mathbf{29.00\%}$ & $+8.15$ \\
\cmidrule(lr){2-15}
& RL vs SFT & SFT
& $66.75\%$ & $86.00\%$ & $80.75\%$ & $28.50\%$ & $22.80\%$ & $17.60\%$ & $48.40\%$ & $33.25\%$ & $37.25\%$ & $33.25\%$ & $24.25\%$ & $-0.41$ \\
& Reward & Cor+Cov-GRPO
& $69.50\%$ & $88.00\%$ & $87.00\%$ & $32.00\%$ & $43.20\%$ & $\mathbf{33.20\%}$ & $52.00\%$ & $39.00\%$ & $43.50\%$ & $37.00\%$ & $29.00\%$ & $+6.37$ \\
& Reward & Cor+Dir-GRPO
& $67.00\%$ & $88.40\%$ & $87.25\%$ & $33.00\%$ & $43.60\%$ & $30.00\%$ & $53.20\%$ & $40.00\%$ & $44.00\%$ & $37.25\%$ & $28.50\%$ & $+6.26$ \\
& Stage & Stage-I CEDAR-GRPO
& $67.50\%$ & $86.00\%$ & $84.25\%$ & $35.50\%$ & $45.20\%$ & $29.20\%$ & $51.20\%$ & $40.25\%$ & $40.50\%$ & $36.75\%$ & $26.00\%$ & $+5.36$ \\
& Stage & Stage-II CEDAR-GRPO
& $68.25\%$ & $86.80\%$ & $85.75\%$ & $\mathbf{37.75\%}$ & $34.40\%$ & $26.80\%$ & $51.20\%$ & $35.00\%$ & $\mathbf{46.25\%}$ & $35.75\%$ & $24.75\%$ & $+4.49$ \\
& Generic reasoning & General Cor-GRPO
& $64.50\%$ & $86.40\%$ & $83.50\%$ & $26.50\%$ & $19.20\%$ & $17.60\%$ & $41.60\%$ & $37.75\%$ & $37.25\%$ & $33.75\%$ & $26.50\%$ & $-0.80$ \\
\midrule
\multirow{9}{2.7cm}{\raggedright DeepSeek-R1-Distill-Qwen-7B} 
& Reference & Base
& $70.25\%$ & $85.20\%$ & $81.50\%$ & $30.00\%$ & $26.40\%$ & $38.40\%$ & $49.60\%$ & $31.50\%$ & $42.50\%$ & $35.75\%$ & $23.75\%$ & $0.00$ \\
& Reference & Cor-GRPO
& $73.50\%$ & $88.00\%$ & $82.25\%$ & $31.50\%$ & $56.40\%$ & $40.40\%$ & $50.00\%$ & $33.00\%$ & $42.25\%$ & $36.25\%$ & $22.75\%$ & $+3.77$ \\
& Main & \textbf{CEDAR-GRPO}
& $\mathbf{78.50\%}$ & $\mathbf{89.60\%}$ & $\mathbf{87.75\%}$ & $35.50\%$ & $\mathbf{57.20\%}$ & $\mathbf{49.60\%}$ & $51.60\%$ & $35.50\%$ & $45.75\%$ & $\mathbf{39.50\%}$ & $24.75\%$ & $+7.31$ \\
\cmidrule(lr){2-15}
& RL vs SFT & SFT
& $72.00\%$ & $86.40\%$ & $80.00\%$ & $34.50\%$ & $29.60\%$ & $36.00\%$ & $47.20\%$ & $33.25\%$ & $38.75\%$ & $32.75\%$ & $19.75\%$ & $-0.42$ \\
& Reward & Cor+Cov-GRPO
& $73.75\%$ & $88.00\%$ & $83.25\%$ & $31.75\%$ & $\mathbf{57.20\%}$ & $41.20\%$ & $50.40\%$ & $33.75\%$ & $43.25\%$ & $36.00\%$ & $23.50\%$ & $+4.29$ \\
& Reward & Cor+Dir-GRPO
& $72.75\%$ & $87.60\%$ & $82.75\%$ & $31.25\%$ & $56.80\%$ & $40.80\%$ & $49.60\%$ & $33.75\%$ & $43.50\%$ & $36.75\%$ & $22.50\%$ & $+3.93$ \\
& Stage & Stage-I CEDAR-GRPO
& $69.50\%$ & $87.20\%$ & $82.75\%$ & $33.25\%$ & $42.00\%$ & $27.60\%$ & $\mathbf{55.20\%}$ & $\mathbf{41.25\%}$ & $42.50\%$ & $35.25\%$ & $\mathbf{29.25\%}$ & $+2.81$ \\
& Stage & Stage-II CEDAR-GRPO
& $72.75\%$ & $88.80\%$ & $86.50\%$ & $\mathbf{40.25\%}$ & $38.40\%$ & $27.20\%$ & $51.60\%$ & $33.25\%$ & $\mathbf{48.00\%}$ & $35.00\%$ & $27.50\%$ & $+3.13$ \\
& Generic reasoning & General Cor-GRPO
& $69.25\%$ & $88.40\%$ & $82.75\%$ & $28.50\%$ & $22.00\%$ & $27.60\%$ & $43.60\%$ & $34.25\%$ & $40.50\%$ & $33.75\%$ & $25.25\%$ & $-1.73$ \\
\bottomrule
\end{tabular}%
}
\caption{Unified ablation results on the two evaluated backbones, isolating the effects of supervised fine-tuning, reward composition, training stages, and generic reasoning post-training. Performance is compared against our main CEDAR-GRPO method and the Cor-GRPO baseline. All columns report task-native metrics on the same held-out evaluation datasets as Table~\ref{tab:main_results}.}
\label{tab:unified_ablations}
\end{table*}

\subsubsection{RL versus supervised fine-tuning on synthetic rationales}
\label{sec:ablation_sft}

This ablation asks whether the gains come from reinforcement learning or simply from additional exposure to the same abductive tasks and answer formats. We construct an SFT dataset from the same training pool used for RL. Because not all original examples include complete reasoning traces, we generate synthetic rationale chains in the same \texttt{<think>} and \texttt{<answer>} format used during RL training; details are given in Appendix~\ref{app:sft_dataset_creation}. We then fine-tune the same base model with the same PEFT configuration and evaluate it on the identical held-out suite.

This creates a matched comparison: SFT sees the same task distribution and response format, but learns by imitating fixed rationales rather than by exploring responses under a reward signal. As shown in Table~\ref{tab:unified_ablations}, SFT is weaker than the closest RL counterpart, Cor-GRPO, which also omits the composite process reward. On Qwen3-4B, Cor-GRPO outperforms SFT on every held-out task; on DeepSeek-R1-Distill-Qwen-7B, it is higher on 9 of 11 tasks and better on average. This suggests that outcome-driven exploration provides benefits beyond simply imitating synthetic rationales.

\subsubsection{Reward composition}
\label{sec:ablation_rewards}

CEDAR-GRPO uses a composite reward that combines deterministic final-answer correctness with evidence coverage and evidence--explanation directionality:
\[
    r = \lambda_{\mathrm{cor}} r_{\mathrm{cor}}
      + \lambda_{\mathrm{cov}} r_{\mathrm{cov}}
      + \lambda_{\mathrm{dir}} r_{\mathrm{dir}} .
\]
To isolate the role of the two process rewards, we keep correctness active and remove one process term at a time. In each two-term variant, the included rewards receive equal weight ($0.5$ each), while the omitted reward is set to zero. Thus, Cor+Cov-GRPO removes directionality, while Cor+Dir-GRPO removes evidence coverage. The ``Reward'' rows in Table~\ref{tab:unified_ablations} compare these variants with Cor-GRPO and the full CEDAR-GRPO objective.

The results show that both process rewards contribute. Adding either one to correctness improves average performance over Cor-GRPO, but neither partial objective matches the full composite reward. On Qwen3-4B, CEDAR-GRPO reaches 52.1 average accuracy, compared with 50.3 for Cor+Cov-GRPO and 50.2 for Cor+Dir-GRPO; on DeepSeek-R1-Distill-Qwen-7B, the corresponding scores are 54.1, 51.1, and 50.7. This suggests that coverage and directionality are complementary: each helps beyond final-answer correctness, but the strongest transfer comes from optimizing them jointly.

\subsubsection{Isolating Stage~I and Stage~II training}
\label{sec:ablation_stages}

Our data collection is built around the claim that abductive reasoning requires both hypothesis generation and hypothesis selection. To test this directly, we train two restricted variants. The Stage~I-only model uses only the hypothesis-generation sources: AbductionRules, Crypto, and List Function. The Stage~II-only model uses only the hypothesis-selection or hypothesis-evaluation sources: UniADILR-HGc, Balanced COPA cause, CauseLogics, and CLIMATE-FEVER. We keep the optimization budget and checkpoint-selection protocol matched as closely as possible to the full mixture.

As shown in the ``Stage'' rows of Table~\ref{tab:unified_ablations}, the restricted variants retain some task-specific strengths, but neither matches the breadth of the full training mixture. On Qwen3-4B, CEDAR-GRPO reaches 52.1 average accuracy, compared with 49.3 for Stage~I-only and 48.4 for Stage~II-only training. The same pattern holds on DeepSeek-R1-Distill-Qwen-7B, where the full mixture reaches 54.1, compared with 49.6 and 49.9 for the two restricted variants. These results suggest that generation- and selection-oriented data provide complementary signals, and that broad transfer depends on training over both stages.

\subsubsection{Training on generic reasoning data}
\label{sec:ablation_generic_reasoning}

Finally, we test whether the observed gains are specific to abductive data or can be obtained by applying GRPO to a generic reasoning mixture. We create a matched post-training condition using deductive and general reasoning datasets with the same backbone, similar number of examples, answer format, and RL budget. The matched general-reasoning mixture is detailed in Appendix~\ref{app:general_reasoning_data}. This controls for improvements from longer reasoning traces, answer-format practice, or generic reinforcement learning on verifiable problems.

As shown in the ``Generic reasoning'' rows of Table~\ref{tab:unified_ablations}, the generic-reasoning control underperforms the matched abductive correctness-only baseline, Cor-GRPO. On Qwen3-4B, Cor-GRPO is higher on every held-out task and averages 49.9 compared with 43.2 for General Cor-GRPO. On DeepSeek-R1-Distill-Qwen-7B, it also performs better on average, 50.6 versus 45.1, with the largest gaps appearing on the MuSR tasks. This suggests that the gains do not come simply from correctness-only GRPO on verifiable reasoning data, but from training on abductive data in particular.

\section{Conclusion}
We introduced CEDAR-GRPO, a composite-reward post-training framework that treats abduction as a general reasoning capability rather than a benchmark-specific skill. Across four open-weight backbones and eleven held-out datasets, CEDAR-GRPO consistently improved performance over both base models and correctness-only GRPO, with gains of up to 30.8 points and transfer to abduction-adjacent settings such as clinical diagnosis, ML debugging, and long-context investigation. Ablations indicate that these gains reflect the abductive training mixture and the composite reward rather than supervised exposure or generic verifiable-reward post-training. Overall, CEDAR-GRPO is a practical step toward more general and reliable abductive reasoning in language models.

\section{Limitations}
Despite the consistent improvements reported above, several limitations remain. First, our evaluation is still constrained by benchmark-style settings and largely closed-form outcomes, and it remains unclear how far these gains extend to truly open-ended or interactive explanatory tasks where the answer space is not predefined. Second, all backbones evaluated are in the 4--8B parameter range, leaving open whether the observed patterns hold at larger scales where base reasoning capabilities and training dynamics may differ substantially. Third, although we complement final-answer accuracy with process-level metrics to validate that the observed improvements reflect genuine changes in reasoning behavior, these metrics are themselves computed by an LLM-as-judge, and since the evidence-coverage and directionality rewards used during training rely on the same kind of judgments, there is a potential circularity between optimization signal and evaluation. Conclusions about reasoning quality would therefore be strengthened by human evaluation, both as a check on the process-level scores on our existing evaluation suite and through dedicated human assessment of model outputs on open-ended generation tasks where automatic metrics are least reliable. Finally, our training pool is relatively small at 2{,}400 instances (Section~\ref{sec:ablation_sft}); while this scale is sufficient to demonstrate the effects studied here, it leaves open how the method behaves under substantially larger or more diverse abductive training data, and we examine the scale question only for the model and not for the data.

\bibliography{references}

\clearpage
\appendix

\section{Additional Dataset Notes for Data Collection}
\label{app:data_collection}

The notes below distinguish between the original benchmark as introduced in its source paper and the way we operationalize it in our study. This distinction matters because several datasets are released as task collections or reasoning-type subsets rather than as fixed train/dev/test corpora, and in several cases our subset choice is narrower than the benchmark as a whole. 

\subsection{Training and validation sources}

\begin{table*}[h!]
\centering
{\footnotesize
\setlength{\tabcolsep}{3.5pt}
\renewcommand{\arraystretch}{1.05}
\begin{tabular}{p{4.2cm}p{2.6cm}cp{5.2cm}}
\toprule
Dataset & Relation to abduction & Samples & Training role \\
\midrule
\makecell[l]{UniADILR-HGc (abductive)\\\citep{sheng-etal-2025-evaluating}}
&
\makecell[l]{Stage II}
& 400
&
Abstract logical abductive selection with distractor premises; compact signal for evaluating candidate explanations. \\
\makecell[l]{Balanced COPA (cause)\\\citep{kavumba-etal-2020-balanced-copa}}
&
\makecell[l]{Stage II}
& 400
&
Commonsense causal selection while reducing superficial lexical cues. \\
\makecell[l]{CauseLogics\\\citep{he2024causejudger}}
&
\makecell[l]{Stage II-adjacent}
& 400
&
Binary candidate--cause verification with longer reasoning chains; deeper evaluative reasoning. \\
\makecell[l]{CLIMATE-FEVER\\\citep{diggelmann2020climatefever}}
&
\makecell[l]{Stage II-adjacent}
& 400
&
Four-way claim--evidence plausibility judgment over supports, refutes, not enough info, and disputed; broader verification-style reasoning. \\
\makecell[l]{AbductionRules\\\citep{young2022abductionrules}}
&
\makecell[l]{Stage I}
& 400
&
Abductive generation of missing explanatory facts from rule contexts. \\
\makecell[l]{Crypto\\\citep{li2025patternsprinciplesfragilityinductive}}
&
\makecell[l]{Stage I}
& 200
&
Cipher-rule learning (Caesar, Atbash) from examples. \\
\makecell[l]{List Function\\\citep{rule2020building}}
&
\makecell[l]{Stage I}
& 200
&
Latent list-transformation inference. \\
\bottomrule
\end{tabular}
\caption{
Training and validation datasets used for GRPO fine-tuning.
Each dataset contributes a fixed number of samples (400 for most datasets, 200 for Crypto and List Function), split 80/20 into train/validation; Which sums up to a total number of 2400 samples, consisting of 1920 training and 480 validation samples.
}
\label{tab:train_data}
}
\end{table*}

\paragraph{UniADILR-HGc (abductive split).}
UniADILR was introduced to study generalization across abductive, deductive, and inductive logical reasoning~\citep{sheng-etal-2025-evaluating}. A typical example contains a target statement, a context with both relevant premises and distractors, and a proof-style justification. In the source benchmark, the program-synthesized PSy portion is large, while the human--GPT-corrected HGc portion is smaller and more realistic. We use 400 randomly sampled examples from the abductive HGc portion and split them into 320 training and 80 validation samples. We use this subset because it provides a compact and relatively clean Stage~II signal.

\paragraph{Balanced COPA (cause split).}
Balanced COPA keeps the COPA format of a premise plus two alternatives, but mirrors the training data so that lexical cues are balanced between correct and incorrect answers~\citep{kavumba-etal-2020-balanced-copa}. The original resource is primarily a rebalanced training set of 1,000 instances, evaluated against the original COPA test set. In our study, we use the training split only, filter examples with \texttt{question == "cause"}, randomly sample 400 examples, and split them into 320 training and 80 validation samples. The cause direction is used because it remains close to canonical commonsense abduction.

\paragraph{CauseLogics.}
CauseLogics is a structured decision benchmark with premises, rules, a phenomenon, a candidate cause, and a binary validity label~\citep{he2024causejudger}. The full benchmark contains four difficulty levels, each corresponding to a different reasoning-chain length. We randomly sample 100 examples from each of the four levels and split each level into 80 training and 20 validation examples, producing 320 training and 80 validation samples overall. This keeps the training pool balanced across reasoning-chain lengths while still exposing the model to harder multi-step cases.

\paragraph{CLIMATE-FEVER.}
CLIMATE-FEVER is a FEVER-style claim-verification dataset built from real-world climate claims and Wikipedia evidence~\citep{diggelmann2020climatefever}. The original release contains 1,535 claims and 7,675 claim--evidence pairs, labeled as supports, refutes, not enough info, or disputed. In our pipeline, it functions as a Stage~II evaluative proxy: the model must assess which evidential relation best characterizes how a piece of evidence bears on a claim. We randomly sample 400 examples from the test split and divide them into 320 training and 80 validation samples.

\paragraph{AbductionRules.}
AbductionRules is a formal abductive reasoning benchmark in which the model is given a rule context together with an unexpected observation and must generate the missing fact that would explain it~\citep{young2022abductionrules}. We construct a balanced subset using the four benchmark variants (\texttt{Abduction-Animal-Simple}, \texttt{Abduction-Animal}, \texttt{Abduction-Person-Simple}, and \texttt{Abduction-Person}). For training, we sample 80 contexts from each variant using \texttt{train.jsonl}, producing 320 training examples in total. For validation, we sample 20 contexts from each variant using \texttt{dev.jsonl}, producing 80 validation examples. One query is sampled per context. We use this dataset as Stage~I training data because the task directly targets hypothesis construction rather than answer selection.

\paragraph{List Function.}
The List Function benchmark is task-based rather than a single canonical corpus~\citep{rule2020building}. Each task provides several input/output demonstrations and asks the model to infer the latent list transformation and apply it to a new input. Under our formalism, that hidden transformation is the missing explanatory rule; when expressed as code in our setup, the generated function serves as an explicit candidate hypothesis about the governing rule. We randomly sample 200 tasks from the full dataset and split them into 160 training and 40 validation tasks. For each task, we use the \texttt{train["normal"]} demonstrations together with the test examples, while ignoring the out-of-distribution and noisy variants. We use this dataset as Stage~I rule-learning data.

\paragraph{Crypto (Caesar and Atbash).}
We draw Crypto examples from the cryptography tasks of the Robust Rule Induction benchmark introduced by Li et al.~\citep{li2025patternsprinciplesfragilityinductive}. In our study, we use only the Caesar and Atbash subsets, sampling 100 examples from each to produce 200 total instances, which are divided into 160 training and 40 validation examples. Each instance requires inferring the latent cipher rule from input--output demonstrations, so we treat the dataset as Stage~I rule-learning data.

\subsection{Evaluation sources}

\begin{table*}[h!]
\centering
{\footnotesize
\setlength{\tabcolsep}{3.2pt}
\renewcommand{\arraystretch}{1.08}
\begin{tabular}{p{3.6cm}p{2.2cm}c p{6.6cm}}
\toprule
Dataset & Relation & Samples & Evaluation role \\
\midrule
\multicolumn{4}{l}{\textbf{Classic abductive benchmarks}} \\
\makecell[l]{ART ($\alpha$NLI)\\\citep{bhagavatula2019abductive-commonsense-reasoning}}
& Stage II
& 400
& Conventional Stage~II abductive selection between alternative hypotheses over narrative bridges. \\
\makecell[l]{NeuLR\\\citep{xu2023logicalreasoners}}
& Stage I
& 400
& Conventional Stage~I abductive split; missing-fact generation in abstract logical contexts. \\
\midrule
\multicolumn{4}{l}{\textbf{Stage~II evidence-update evaluation}} \\
\makecell[l]{Defeasible NLI\\\citep{rudinger2020thinking-like-a-skeptic}}
& Stage II-adjacent
& 400
& Determines whether an update strengthens or weakens a hypothesis; tests evidence-based plausibility assessment. \\
\midrule
\multicolumn{4}{l}{\textbf{Abductive interpretations in neighboring domains}} \\
\makecell[l]{GoEmotions\\\citep{demszky-etal-2020-goemotions}}
& Domain-abductive
& 400
& Infers latent emotion as the explanation for an observed utterance. \\
\makecell[l]{MuSR: Murder\\\citep{sprague2023musr}}
& Domain-abductive
& 250
& Detective culprit inference; infers the most plausible explanation for a murder. \\
\makecell[l]{MedQA\\\citep{jin2020disease}}
& Domain-abductive
& 400
& Clinical diagnosis: infers the most plausible illness, problem, or mechanism from patient background and symptoms. \\
\makecell[l]{ML-debugging\\\citep{huang-etal-2025-mldebugging}}
& Domain-abductive
& 400
& Debugging as inferring the cause of an observed error or malfunction and producing a fix. \\
\midrule
\multicolumn{4}{l}{\textbf{Non-abductive controls}} \\
\makecell[l]{Balanced COPA (effect)\\\citep{kavumba-etal-2020-balanced-copa}}
& Deductive control
& 250
& Tests forward causal reasoning in the effect direction. \\
\makecell[l]{MuSR: Object\\\citep{sprague2023musr}}
& Multistep control
& 250
& Tests complex long-context reasoning over object locations. \\
\makecell[l]{MuSR: Team\\\citep{sprague2023musr}}
& Multistep control
& 250
& Tests complex long-context reasoning over team assignments. \\
\makecell[l]{StrategyQA\\\citep{geva-etal-2021-aristotle}}
& General control
& 400
& Broad reasoning benchmark not explicitly framed as abduction. \\
\midrule
\multicolumn{2}{l}{\textbf{Total}} & \textbf{3,800} & \\
\bottomrule
\end{tabular}
\caption{
Held-out evaluation datasets used for model assessment.
These datasets are not used for GRPO training or validation-based selection.
The Samples column reports the number of held-out examples evaluated in this study, including full selected subsets for some datasets and fixed-size samples from larger source splits for others.
}
\label{tab:eval_data}
}
\end{table*}

\paragraph{ART ($\alpha$NLI).}
ART is a canonical abductive commonsense benchmark~\citep{bhagavatula2019abductive-commonsense-reasoning}. In its $\alpha$NLI formulation, the model is given two observations and must choose which of two candidate hypotheses best bridges them. The same paper also introduces $\alpha$NLG for generation, but here we use only $\alpha$NLI, since our goal on this benchmark is a clean held-out test of conventional Stage~II hypothesis selection.

\paragraph{NeuLR.}
NeuLR is designed to remove topical and commonsense cues so that logical form is tested in a content-neutral setting~\citep{xu2023logicalreasoners}. The benchmark spans deductive, inductive, and abductive reasoning over abstract facts and rules. We use 400 examples from its abductive split because this subset provides the clearest held-out test of conventional Stage~I missing-fact generation rather than surface-level plausibility scoring.

\paragraph{Defeasible NLI (DefNLI).}
Defeasible NLI extends several existing inference resources with a defeasible-update formulation~\citep{rudinger2020thinking-like-a-skeptic}. The classification task takes a premise/hypothesis pair together with an update and asks whether the update strengthens or weakens the hypothesis. We use this classification view because it directly probes the evaluative side of Stage~II reasoning, in a role analogous to CLIMATE-FEVER in the training pool: the model must judge how new evidence changes the plausibility of a candidate conclusion or explanation.

\paragraph{GoEmotions (GoEmo).}
GoEmotions is a large multi-label emotion dataset of Reddit comments annotated with 27 emotion categories plus \emph{Neutral}~\citep{demszky-etal-2020-goemotions}. The benchmark is not an abductive dataset in the standard sense, but it can still be read as a domain-specific hidden-state inference problem: from an observed utterance, the model must infer the latent affective state that best explains it. We use it precisely for this out-of-domain abductive interpretation.

\paragraph{MuSR: Murder (MuSR-M).}
MuSR is a long-context multiple-choice benchmark with three narrative domains: murder mysteries, object placements, and team allocations~\citep{sprague2023musr}. The murder split is the most directly compatible with an abductive interpretation, since the task is to infer the culprit or explanatory sequence of events that best accounts for the available evidence. We therefore treat MuSR-M as a domain-shifted abductive benchmark.

\paragraph{MuSR: Object and MuSR: Team (MuSR-O/T).}
The object-placement and team-allocation splits of MuSR are less cleanly abductive than the murder split. They still require long-context, multi-step reasoning over dispersed constraints, but the target is better understood as recovering an object location or assignment structure rather than explaining an observed outcome. We therefore use MuSR-O and MuSR-T as non-abductive multistep controls.

\paragraph{MedQA.}
MedQA is a multiple-choice question-answering benchmark built around complex multi-hop medical reasoning~\citep{jin2020disease}. Each question is accompanied in the original release by a large-scale collection of medical textbook paragraphs for evidence retrieval, but the standard prediction target is the final multiple-choice answer. We use it as a specialized domain-abductive benchmark because many instances require inferring the most plausible illness, clinical problem, treatment, or underlying mechanism from patient background information, symptoms, and test results.

\paragraph{ML-debugging.}
ML-debugging is a code debugging benchmark built around identifying and resolving errors in multi-library Python scenarios~\citep{huang-etal-2025-mldebugging}. Each buggy code snippet is accompanied in the original release by execution feedback and test cases, but the standard prediction target is the final repaired code sequence. We use it as a specialized domain-abductive benchmark because debugging requires tracing an observed error, failed test, or malfunction back to its likely cause and then producing a repair.

\paragraph{Balanced COPA (effect split).}
We use the effect direction of Balanced COPA only at evaluation time~\citep{kavumba-etal-2020-balanced-copa}. This split keeps the same two-choice causal format as the cause direction used in training, but reverses the direction of reasoning. We treat it as a non-abductive deductive-style control because the model is asked to reason forward from a premise toward a likely effect, rather than backward from an observation toward an explanation.

\paragraph{StrategyQA (StratQA).}
StrategyQA is a yes/no question-answering benchmark built around implicit multi-step reasoning~\citep{geva-etal-2021-aristotle}. Each question is accompanied in the original release by intermediate decompositions and supporting evidence, but the standard prediction target is simply the final yes/no answer. We do not treat it as an abductive benchmark. Instead, it functions as a non-abductive general-reasoning control that helps us determine whether any observed gains are specific to explanation-centered reasoning or reflect broader changes in reasoning behavior.

Taken together, these choices create an intentional gradient: classic abduction in ART and NeuLR, Stage~II evidence-update evaluation in Defeasible NLI, abductive interpretations in neighboring domains through GoEmotions, MuSR-M, MedQA, and ML-debugging, and explicit non-abductive controls through Balanced COPA-effect, MuSR-O/T, and StrategyQA. This structure supports a more precise interpretation of results: improvements on ART, NeuLR, and Defeasible NLI would indicate more direct abductive transfer, improvements on the domain-abductive group would suggest transfer to explanation-centered reasoning outside standard abduction benchmarks, and improvements on the controls would indicate broader or less specific changes in the model's reasoning profile.

\section{Prompts Used in Training and Evaluation} 
\label{app:prompts}

\subsection{Prompts Used in Training}

\subsubsection{Prompt Template: UniADILR}

\noindent\textbf{System Prompt:}
\begin{lstlisting}[breaklines=true, basicstyle=\small\ttfamily]
You are an expert in logical reasoning and abductive inference. Your task is to identify which sentences from a given context provide the necessary evidence to support or explain a hypothesis.

You will be provided with:
1. A Context containing multiple numbered sentences (sent1, sent2, sent3, etc.)
2. A Hypothesis that needs to be supported or explained

Your goal is to identify which sentence(s) from the context, when combined, provide the logical foundation for the hypothesis through abductive reasoning.

## Instructions:
1. Carefully read all sentences in the context
2. Analyze the hypothesis
3. Identify which sentences, when combined, best explain or support the hypothesis
4. Consider both direct evidence and logical connections
5. Think step by step.

## Output Format:
You MUST provide your answer in the following format:

<think>
[Think step by step here]
</think>
<answer>
[Sentence numbers only, comma-separated. For example: 5, 13 or 2, 7, 9]
</answer>

CRITICAL: The answer section must contain ONLY the sentence numbers separated by commas. Do not include the word "sent" or any other text.
\end{lstlisting}

\vspace{0.8em}

\noindent\textbf{User Prompt:}
\begin{lstlisting}[breaklines=true, basicstyle=\small\ttfamily]
Context:
{context_str}

Hypothesis:
{hypothesis}

Which sentence numbers provide the necessary evidence for the hypothesis?
\end{lstlisting}

\subsubsection{Prompt Template: Copa (cause)}

\noindent\textbf{System Prompt:}
\begin{lstlisting}[breaklines=true, basicstyle=\small\ttfamily]
You are an expert in logical reasoning and abductive inference. Your task is to determine which of two given choices represents the most plausible cause for a given premise.

You will be provided with:
1. A Premise describing a situation or event
2. Two Choices (Choice 1 and Choice 2)

Your goal is to select the choice that best explains WHY the premise happened - identifying the root cause that led to the described situation.

## Instructions:
1. Carefully read the premise
2. Think step by step to evaluate both choices as potential causes
3. Consider common sense, real-world knowledge, and typical causal relationships when making your decision
4. Select the choice that represents the most plausible and direct cause
5. Think step by step.

## Output Format:
You MUST provide your answer in the following format:

<think>
[Think step by step here]
</think>
<answer>
[Either "1" or "2" - just the number, nothing else]
</answer>

CRITICAL: The answer section must contain ONLY the number 1 or 2. Do not include any other text, explanation, or punctuation.
\end{lstlisting}

\vspace{0.8em}

\noindent\textbf{User Prompt:}
\begin{lstlisting}[breaklines=true, basicstyle=\small\ttfamily]
Premise: {example['premise']}

Choice 1: {example['choice1']}
Choice 2: {example['choice2']}

Which choice is the most plausible cause for the premise?
\end{lstlisting}

\subsubsection{Prompt Template: CauseLogics}

\noindent\textbf{System Prompt:}
\begin{lstlisting}[breaklines=true, basicstyle=\small\ttfamily]
You are an expert logician and careful reasoning assistant. Your task is to identify whether a given Possible Cause, when added to the provided knowledge base, logically entails an observed Phenomenon.

You will be provided with:
1. A set of Premises (facts)
2. A set of Rules (implications)
3. An observed Phenomenon
4. A Possible Cause (a hypothesis)

Your goal is to determine whether the Phenomenon can be logically inferred by forward reasoning using ONLY the given Premises + Rules (+ the Possible Cause).

## Instructions:
1. Carefully read all Premises and Rules
2. Assume the Possible Cause is added as an additional premise
3. Using ONLY the given Premises + Rules (+ the Possible Cause), reason forward
4. Decide whether the Phenomenon can be logically inferred
    - If the Phenomenon can be inferred, the Possible Cause is TRUE
    - If the Phenomenon cannot be inferred, the Possible Cause is FALSE
5. Think step by step.

## Output Format:
You MUST provide your answer in the following format:

<think>
[Think step by step here]
</think>
<answer>
[Output exactly one of these two options: TRUE, FALSE]
</answer>

CRITICAL: The answer section must contain ONLY one of these two options: TRUE or FALSE. Do not include any other text.
\end{lstlisting}

\vspace{0.8em}

\noindent\textbf{User Prompt:}
\begin{lstlisting}[breaklines=true, basicstyle=\small\ttfamily]
Premises:
{premises_text}

Rules:
{rules_text}

Phenomenon:
{str(phenomenon)}

Possible Cause:
{str(possible_cause)}

Is the Possible Cause logically TRUE or FALSE?
\end{lstlisting}

\subsubsection{Prompt Template: Climate Fever}

\noindent\textbf{System Prompt:}
\begin{lstlisting}[breaklines=true, basicstyle=\small\ttfamily]
You are an expert climate scientist and professional fact-checker. Your task is to determine whether a set of provided evidences supports, refutes, disputed or is insufficient to evaluate a specific claim.

You will be provided with:
1. A specific Claim
2. A list of Evidences

Your goal is to decide whether the Evidence SUPPORTS or REFUTES or DISPUTED the Claim, or if there is NOT ENOUGH INFO, and to justify that decision by citing specific parts of the evidence.

## Instructions:
1. Carefully read the Claim and all provided Evidences
2. Determine if the Evidence SUPPORTS or REFUTES or DISPUTED the Claim, or if there is NOT ENOUGH INFO
3. Think step by step about how the specific parts of the evidence relate to the claim
4. Output the final label
5. Think step by step.

## Output Format:
You MUST provide your answer in the following format:

<think>
[Think step by step here]
</think>

<answer>
[Output exactly one of these four options: SUPPORTS, REFUTES, DISPUTED, NOT ENOUGH INFO]
</answer>

CRITICAL: The answer section must contain ONLY one of these four options: SUPPORTS, REFUTES, DISPUTED, NOT ENOUGH INFO. Do not include any other text.
\end{lstlisting}

\vspace{0.8em}

\noindent\textbf{User Prompt:}
\begin{lstlisting}[breaklines=true, basicstyle=\small\ttfamily]
Claim:
{claim}

Evidence:
{evidence_text}

Does the provided evidence SUPPORT, REFUTE, DISPUTED or provide NOT ENOUGH INFO for the claim?
\end{lstlisting}

\subsubsection{Prompt Template: Abduction Rules}

\noindent\textbf{System Prompt:}
\begin{lstlisting}[breaklines=true, basicstyle=\small\ttfamily]
You are an expert in logical reasoning and abductive inference. Your task is to identify the single missing fact that, when added to a given context, makes a query logically decidable.

You will be provided with:
1. A Context containing facts and rules
2. A Query that is currently not decidable from the context alone

Your goal is to infer ONE additional fact that, when combined with the context, allows the query to be either:
- proved true, or
- proved false

## Instructions:
1. Carefully read all facts and rules in the context
2. Analyze the query
3. Identify the single missing fact that would make the query decidable
4. Prefer a direct, minimal explanation:
    - Output exactly one fact
    - Do not output a rule
    - Do not output multiple facts
    - Do not paraphrase beyond the style already used in the context
5. The fact should be one that works with the existing rules and facts to prove or disprove the query
6. Be careful with negation:
    - Sometimes the right missing fact helps prove the query
    - Sometimes it helps derive the opposite of the query, thereby disproving it

## Output Format:
You MUST provide your answer in the following format:

<think>
[Explain your thought process: which rule(s) matter, which existing facts are relevant, and why the missing fact makes the query provable or disprovable]
</think>

<answer>
[Output the single missing fact only, exactly as a natural-language sentence ending with a period]
</answer>

CRITICAL:
- The answer section must contain ONLY one missing fact.
- Do not include any extra commentary in the answer section.
- Do not output more than one sentence.
- Do not output a rule; output a fact about an entity in the context.
\end{lstlisting}

\vspace{0.8em}

\noindent\textbf{User Prompt:}
\begin{lstlisting}[breaklines=true, basicstyle=\small\ttfamily]
Context:
{context}

Query:
{query}

Based on the context and query above, identify the single missing fact that, when added to the context, makes the query logically decidable.
\end{lstlisting}

\subsubsection{Prompt Template: Crypto}

\noindent\textbf{System Prompt:}
\begin{lstlisting}[breaklines=true, basicstyle=\small\ttfamily]
You are an expert at inferring exact string transformation rules from examples and expressing them as correct Python functions.

You will be given several training examples. Each example contains:
- Input: a string
- Output: the result of applying the same hidden transformation rule to the input

Infer the transformation rule that is consistent with ALL training examples, then write a general Python implementation of that rule.

Before answering, make sure the same rule explains all examples exactly and consistently at the character level.
Think abductively: consider alternative hypotheses and choose the one that explains all examples exactly.

Output format (MUST follow exactly):
<think>
[Explain your thought process: reason step by step about the possible rules, consider alternative hypotheses, and explain why the rule you chose best fits all examples.]
</think>
<answer>
def transform(s):
    ...
</answer>

Code requirements:
- Define EXACTLY one function named transform.
- The function takes one argument: s (a string).
- It MUST return a string.
- NO IMPORTS allowed.
- NO printing, no input(), no randomness.
- Do not hardcode specific training inputs/outputs; generalize the logic.
- Preserve the behavior implied by the examples for all characters that appear.

STRICT FORMATTING RULES:
- Do NOT use markdown code blocks (like ```python) inside the <answer> tags. Just write raw code.
- Do NOT repeat the code. Write the function exactly once.
- Ensure you close the tag with </answer>.
- The <answer> tag must contain ONLY valid Python code, no comments or explanations outside the function.
- Do NOT write any text before <think> or after </answer>.
\end{lstlisting}

\vspace{0.8em}

\noindent\textbf{User Prompt:}
\begin{lstlisting}[breaklines=true, basicstyle=\small\ttfamily]
Training examples:
{train_prompt}

Infer the underlying string transformation and provide the Python function implementation in the required format.
\end{lstlisting}

\subsubsection{Prompt Template: List Function}

\noindent\textbf{System Prompt:}
\begin{lstlisting}[breaklines=true, basicstyle=\small\ttfamily]
You are an expert at inferring simple list transformations from examples and expressing them as correct Python functions.

You will be given several training examples. Each example contains:
- Input:  a list of integers
- Output: the result of applying the same hidden transformation rule to the input

Infer the transformation rule that is consistent with ALL training examples, then write a general Python implementation of that rule.

Output format (MUST follow exactly):
<think>
[Explain your thought process: reason step by step about the possible rules, consider alternative hypotheses, and explain why your final rule best fits ALL training examples.]
</think>
<answer>
def transform(lst):
    ...
</answer>

Code requirements:
- Define EXACTLY one function named transform.
- The function takes one argument: lst (a list of integers).
- It MUST return a list of integers. If the rule results in a single value, return it as a single-element list (e.g., [val]).
- NO IMPORTS allowed.
- NO printing, no input(), no randomness.
- Do not hardcode specific training inputs/outputs; generalize the logic.
- BE ROBUST: Handle edge cases like empty lists or lists with only 1 element.

STRICT FORMATTING RULES:
- Do NOT use markdown code blocks (like ```python) inside the <answer> tags. Just write raw code.
- Do NOT repeat the code. Write the function exactly once.
- Ensure you close the tag with </answer>.
- The <answer> tag must contain ONLY valid Python code, no comments or explanations outside the function.
- Do NOT write any text before <think> or after </answer>.
\end{lstlisting}

\vspace{0.8em}

\noindent\textbf{User Prompt:}
\begin{lstlisting}[breaklines=true, basicstyle=\small\ttfamily]
Training examples:
{train_prompt}

Infer the underlying list transformation and provide the Python function implementation in the required format.
\end{lstlisting}

\subsection{Prompts Used in Evaluation}

\subsubsection{Prompt Template: Art}

\noindent\textbf{System Prompt:}
\begin{lstlisting}[breaklines=true, basicstyle=\small\ttfamily]
You are an expert in abductive reasoning and narrative comprehension. Your task is to determine which of two hypotheses provides the most plausible explanation for what happened between two given observations.

You will be provided with:
1. Observation 1 (the initial situation or event)
2. Observation 2 (the subsequent outcome or resulting event)
3. Two Hypotheses (Hypothesis 1 and Hypothesis 2)

Your goal is to select the hypothesis that logically and narratively bridges the gap between Observation 1 and Observation 2, explaining how the situation transitioned from the first observation to the second.

## Instructions:
1. Carefully read Observation 1 and Observation 2 to understand the chronological and narrative context
2. Evaluate both Hypothesis 1 and Hypothesis 2 as potential bridging events
3. Consider common sense, cause-and-effect relationships, and everyday plausibility
4. Select the hypothesis that best explains the transition
5. Think step by step.

## Output Format:
You MUST provide your answer in the following format:

<think>
[Think step by step here]
</think>
<answer>
[Either "1" or "2" - just the number, nothing else]
</answer>

CRITICAL: The answer section must contain ONLY the number 1 or 2. Do not include any other text, explanation, or punctuation.
\end{lstlisting}

\vspace{0.8em}

\noindent\textbf{User Prompt:}
\begin{lstlisting}[breaklines=true, basicstyle=\small\ttfamily]
Observation 1: {obs1}
Observation 2: {obs2}

Hypothesis 1: {hyp1}
Hypothesis 2: {hyp2}

Which hypothesis better explains the transition from Observation 1 to Observation 2?
\end{lstlisting}

\subsubsection{Prompt Template: Copa (effect)}

\noindent\textbf{System Prompt:}
\begin{lstlisting}[breaklines=true, basicstyle=\small\ttfamily]
You are an expert in logical reasoning and common-sense causal inference. Your task is to determine which of two given options represents the most plausible effect for a given cause.

You will be provided with:
1. A Cause describing a situation or event
2. Two Options (Option 1 and Option 2)

Your goal is to select the option that best describes the direct effect, logical consequence, or most likely resulting action of the given cause.

## Instructions:
1. Carefully read the provided cause
2. Evaluate both Option 1 and Option 2 as potential effects or consequences
3. Consider common sense, real-world knowledge, and typical cause-and-effect relationships
4. Select the option that represents the most plausible direct effect
5. Think step by step.

## Output Format:
You MUST provide your answer in the following format:

<think>
[Think step by step here]
</think>
<answer>
[Either "1" or "2" - just the number, nothing else]
</answer>

CRITICAL: The answer section must contain ONLY the number 1 or 2. Do not include any other text, explanation, or punctuation.
\end{lstlisting}

\vspace{0.8em}

\noindent\textbf{User Prompt:}
\begin{lstlisting}[breaklines=true, basicstyle=\small\ttfamily]
Cause: {premise}

Option 1: {choice1}
Option 2: {choice2}

Which of the following is the most plausible EFFECT of this cause?
\end{lstlisting}

\subsubsection{Prompt Template: DefasibleNLI}

\noindent\textbf{System Prompt:}
\begin{lstlisting}[breaklines=true, basicstyle=\small\ttfamily]
You are an expert in defeasible reasoning and logical analysis. Your task is to determine how new information affects the likelihood of a given hypothesis.

You will be provided with:
1. A Hypothesis (a tentative conclusion)
2. An Update (new information)
3. A Premise (optional contextual background)

Your goal is to analyze the context and decide if the new Update makes the Hypothesis more likely or less likely to be true.

## Instructions:
1. Read the Hypothesis and the Premise (if provided) to understand the initial situation
2. Carefully evaluate the new Update
3. Determine if the Update provides evidence that supports the Hypothesis (strengthens it) or contradicts it (weakens it)
4. Classify the effect as either STRENGTHENS or WEAKENS
5. Think step by step.

## Output Format:
You MUST provide your answer in the following format:

<think>
[Think step by step here]
</think>
<answer>
[STRENGTHENS or WEAKENS]
</answer>

CRITICAL: The answer section must contain ONLY the exact word STRENGTHENS or WEAKENS. Do not include any other text, explanation, or punctuation.
\end{lstlisting}

\vspace{0.8em}

\noindent\textbf{User Prompt:}
\begin{lstlisting}[breaklines=true, basicstyle=\small\ttfamily]
Premise:
{premise}

Hypothesis:
{hypothesis}

Update:
{update}

Does this Update STRENGTHEN or WEAKEN the Hypothesis?
\end{lstlisting}

\subsubsection{Prompt Template: GoEmotion}

\noindent\textbf{System Prompt:}
\begin{lstlisting}[breaklines=true, basicstyle=\small\ttfamily]
You are an expert text analyst and emotion classifier. Your task is to identify all emotions expressed in a given text.

You will be provided with:
1. A short Text to analyze

Your goal is to detect the presence of specific emotions from the following predefined list:
[{_emotions_list_str}]

## Instructions:
1. Carefully read the provided text
2. Analyze the context, tone, and nuance to understand the underlying feelings
3. Match the expressed feelings strictly against the predefined list of available emotions
4. Identify all applicable emotions (use "neutral" if no specific emotion is strongly expressed)
5. Think step by step.

## Output Format:
You MUST provide your answer in the following format:

<think>
[Think step by step here]
</think>
<answer>
[Comma-separated list of applicable emotions]
</answer>

CRITICAL: The answer section must contain ONLY the exact emotion names from the available list, separated by commas if there are multiple (e.g., joy, surprise). Do not include any other text, explanation, or capitalization.
\end{lstlisting}

\vspace{0.8em}

\noindent\textbf{User Prompt:}
\begin{lstlisting}[breaklines=true, basicstyle=\small\ttfamily]
Text: "{text}"

What emotion(s) are expressed in this text?
\end{lstlisting}

\subsubsection{Prompt Template: MedQA}

\noindent\textbf{System Prompt:}
\begin{lstlisting}[breaklines=true, basicstyle=\small\ttfamily]
You are an expert medical clinician and diagnostician. Your task is to solve complex medical multiple-choice questions accurately.

You will be provided with:
1. A medical Problem, which typically includes a clinical vignette or medical question along with four candidate choices (A, B, C, D)

Your goal is to evaluate the clinical presentation and select the single most accurate answer.

## Instructions:
1. Carefully read the medical problem, noting key patient demographics, symptoms, physical exam findings, and lab values where applicable
2. Identify the core medical question being asked (e.g., next best step in management, most likely diagnosis, underlying mechanism)
3. Evaluate all four candidate options (A, B, C, D) using evidence-based clinical reasoning
4. Select the letter corresponding to the correct medical answer
5. Think step by step.

## Output Format:
You MUST provide your answer in the following format:

<think>
[Think step by step here]
</think>
<answer>
[Exactly one letter: A, B, C, or D]
</answer>

CRITICAL: The answer section must contain ONLY the single uppercase letter of the correct choice (A, B, C, or D). Do not include parentheses, periods, or any textual explanation.
\end{lstlisting}

\vspace{0.8em}

\noindent\textbf{User Prompt:}
\begin{lstlisting}[breaklines=true, basicstyle=\small\ttfamily]
Problem: {problem}

Which option is the correct answer?
\end{lstlisting}

\subsubsection{Prompt Template: Musr Murder}

\noindent\textbf{System Prompt:}
\begin{lstlisting}[breaklines=true, basicstyle=\small\ttfamily]
You are a brilliant detective and an expert in deductive reasoning. Your task is to analyze clues to solve complex mysteries.

You will be provided with:
1. Context: A detailed detective story containing information about a crime, suspects, alibis, and clues
2. Problem: A question about the mystery, followed by a list of numbered multiple-choice options

Your goal is to logically deduce the truth from the context and identify the correct choice by its index number.

## Instructions:
1. Carefully read the Context to identify timelines, motives, means, and logical inconsistencies among the suspects' statements
2. Evaluate the Problem and all the provided choices
3. Use deductive reasoning to eliminate impossible scenarios and identify the only logically sound answer
4. Note the index number (e.g., 0, 1, 2, ...) of the correct choice
5. Think step by step.

## Output Format:
You MUST provide your answer in the following format:

<think>
[Think step by step here]
</think>
<answer>
[Exactly one integer representing the index of the correct choice]
</answer>

CRITICAL: The answer section must contain ONLY the numeric index number of the correct choice. Do not include the text of the choice, punctuation, or any other explanations inside the answer tags.
\end{lstlisting}

\vspace{0.8em}

\noindent\textbf{User Prompt:}
\begin{lstlisting}[breaklines=true, basicstyle=\small\ttfamily]
Context:
{context}

Problem:
{problem}

What is the index number of the correct choice?
\end{lstlisting}

\subsubsection{Prompt Template: Musr Team}

\noindent\textbf{System Prompt:}
\begin{lstlisting}[breaklines=true, basicstyle=\small\ttfamily]
You are an expert logical reasoner specializing in evaluating team skills and assigning people to tasks optimally. Your task is to analyze a story describing people, their abilities, and their teamwork dynamics in order to determine the best assignment of people to tasks.

You will be provided with:
1. Context: A story describing several people, their abilities at different tasks, and how well they work with others
2. Problem: A question asking which assignment of people to tasks results in the most effective completion of the tasks, along with multiple-choice options indexed as 0, 1, 2, ...

Your goal is to determine which assignment best utilizes each person's skills while also considering teamwork effectiveness when two people must work together on a task.

## Instructions:
1. Carefully read the Context and identify each person's skill level for the relevant tasks (e.g., great, acceptable, or bad)
2. Determine how well different pairs of people work together when assigned to the same task
3. Remember that one task will require two people working together
4. Consider that if one person is bad at a task, the other person's skill may not fully compensate unless they work well together
5. Evaluate the overall effectiveness of each assignment option
6. Select the option that results in the most effective overall completion of all tasks
7. Think step by step.

## Output Format:
You MUST provide your answer in the following format:

<think>
[Think step by step here]
</think>
<answer>
[Exactly one integer representing the index of the correct choice]
</answer>

CRITICAL: The answer section must contain ONLY the numeric index number of the correct choice. Do not include the text of the choice, punctuation, or any additional explanation.
\end{lstlisting}

\vspace{0.8em}

\noindent\textbf{User Prompt:}
\begin{lstlisting}[breaklines=true, basicstyle=\small\ttfamily]
Context:
{context}

Problem:
{problem}

What is the index number of the correct choice?
\end{lstlisting}

\subsubsection{Prompt Template: Musr Object}

\noindent\textbf{System Prompt:}
\begin{lstlisting}[breaklines=true, basicstyle=\small\ttfamily]
You are an expert logical reasoner specializing in tracking beliefs and object locations in narrative stories. Your task is to analyze a story and determine where a character believes an object is located.

You will be provided with:
1. Context: A story describing characters, their actions, and movements of objects
2. Problem: A question about where a specific character believes an object is located, along with multiple-choice options indexed as 0, 1, 2, ...

Your goal is to determine the correct answer by reasoning about what the character observed and therefore believes about the object's location.

## Instructions:
1. Carefully read the Context and track the object's location throughout the story
2. Track what each character observes when the object is moved
3. If a character observes the object moving, they update their belief about the object's location
4. If a character does NOT observe the object moving (e.g., they are absent or distracted), they will continue to believe the object remains in the last location where they saw it
5. Analyze the Problem and evaluate all provided choices
6. Determine which option correctly represents the character's belief about the object's location
7. Think step by step.

## Output Format:
You MUST provide your answer in the following format:

<think>
[Think step by step here]
</think>
<answer>
[Exactly one integer representing the index of the correct choice]
</answer>

CRITICAL: The answer section must contain ONLY the numeric index number of the correct choice. Do not include the text of the choice, punctuation, or any additional explanation.
\end{lstlisting}

\vspace{0.8em}

\noindent\textbf{User Prompt:}
\begin{lstlisting}[breaklines=true, basicstyle=\small\ttfamily]
Context:
{context}

Problem:
{problem}

What is the index number of the correct choice?
\end{lstlisting}

\subsubsection{Prompt Template: NeuLR Abductive}

\noindent\textbf{System Prompt:}
\begin{lstlisting}[breaklines=true, basicstyle=\small\ttfamily]
You are an expert Forensic Logic Analyst and deductive reasoning specialist. Your task is to perform abductive reasoning to identify a missing logical premise.

You will be provided with:
1. Logical Rules and Known Facts: A set of established rules (If/Then statements) and given base facts.
2. Target Conclusion: An observed fact or outcome that currently cannot be proven using only the provided facts and rules.

Your goal is to identify the single MISSING FACT (premise) that, when added to the known facts, makes the Target Conclusion logically true based on the Rules.

## Instructions:
1. Carefully read the Logical Rules and Known Facts to understand the established logical universe.
2. Analyze the Target Conclusion that needs to be proven.
3. Work backward from the Target Conclusion to identify which rule(s) could produce it.
4. Check the conditions for those rule(s) against the Known Facts.
5. Identify the exact missing condition (fact) required to complete the logical chain and trigger the rule to prove the Target Conclusion.
6. Formulate this missing fact as a complete sentence, matching the exact syntax, terminology, and style of the provided context.
7. Think step by step.

## Output Format:
You MUST provide your answer in the following format:

<think>
[Think step by step here]
</think>
<answer>
[The exact missing fact written as a complete sentence]
</answer>

CRITICAL: The answer section must contain ONLY the missing fact as a single complete sentence (e.g., "NPsw0v0k is ADP37scy8."). Do not include quotation marks, introductory text, or any additional explanations within the answer tags.
\end{lstlisting}

\vspace{0.8em}

\noindent\textbf{User Prompt:}
\begin{lstlisting}[breaklines=true, basicstyle=\small\ttfamily]
Logical Rules and Known Facts:
{rules_block}

Target Conclusion:
{target_fact}

What missing fact is required to conclude the Target Conclusion?
\end{lstlisting}

\subsubsection{Prompt Template: StrategyQA}

\noindent\textbf{System Prompt:}
\begin{lstlisting}[breaklines=true, basicstyle=\small\ttfamily]
You are an expert deductive reasoner and fact-checker. Your task is to answer a yes/no question using the provided evidence.

You will be provided with:
1. A Question: A specific query requiring a YES or NO answer.
2. Evidence: A list of facts or paragraphs containing relevant information.

Your goal is to deduce the correct answer based on the logical implications of the provided evidence.

## Instructions:
1. Carefully read the Question to understand what is being asked.
2. Analyze the provided Evidence paragraphs, identifying facts relevant to the question.
3. Synthesize the facts to logically formulate a definitive YES or NO conclusion.
4. Think step by step.

## Output Format:
You MUST provide your answer in the following format:

<think>
[Think step by step here]
</think>
<answer>
[Output exactly YES or NO]
</answer>

CRITICAL: The answer section must contain ONLY the word YES or the word NO. Do not include any other text, punctuation, or explanations.
\end{lstlisting}

\vspace{0.8em}

\noindent\textbf{User Prompt:}
\begin{lstlisting}[breaklines=true, basicstyle=\small\ttfamily]
Question:
{question}

Evidence:
{evidence_text}

Is the answer to the question YES or NO?
\end{lstlisting}

\subsubsection{Prompt Template: MLDebugging}

\noindent\textbf{System Prompt:}
\begin{lstlisting}[breaklines=true, basicstyle=\small\ttfamily]
You are an expert Python developer and debugger. Your task is to identify and fix errors in Python code snippets.

You will be provided with:
1. Task Instructions: The intended behavior and requirements for the code
2. Buggy Code: The incorrect Python code snippet that is failing its tests
3. Runtime Error / Test Feedback: The execution logs, tracebacks, or failing test results

Your goal is to analyze the failure, correct the bug, and provide the complete, working Python code.

## Instructions:
1. Carefully read the Task Instructions to understand the desired functionality
2. Analyze the Buggy Code together with the Runtime Error / Test Feedback to identify the root cause
3. Determine the minimal necessary fix that preserves correct existing behavior
4. Output a fully self-contained corrected Python solution (no placeholders, no omissions)
5. Think step by step.

## Output Format:
You MUST provide your answer in the following format:

<think>
[Think step by step here]
</think>
<answer>
```python
[Your full, corrected Python code here]
```

</answer>

CRITICAL: The answer section must contain ONLY the full corrected Python code block. No explanations, no extra text, and no additional formatting outside the code block.
\end{lstlisting}

\vspace{0.8em}

\noindent\textbf{User Prompt:}
\begin{lstlisting}[breaklines=true, basicstyle=\small\ttfamily]
Task Instructions:
{instruct_prompt}

Buggy Code:

```python
{bug_code}
```

Runtime Error / Test Feedback:
{runtime_feedback}

What is the fully corrected Python code?
\end{lstlisting}

\section{Training and Implementation Details}
\label{sec:appendix_implementation}

\subsection{Hyperparameters and Model Specifications}

The training pipeline uses Unsloth for memory-efficient post-training of \texttt{DeepSeek-R1-Distill-Qwen-7B}, \texttt{Qwen3-4B}, \texttt{Qwen3-8B}, and \texttt{Llama-3.1-8B-Instruct}. All models use 4-bit NormalFloat (NF4) quantization and Low-Rank Adaptation (LoRA). Table~\ref{tab:hyperparams} reports the shared configuration.

\begin{table*}[!t]
\centering
\small
\begin{tabularx}{\textwidth}{l X l X}
\toprule
\textbf{Hyperparameter} & \textbf{Value} & \textbf{Hyperparameter} & \textbf{Value} \\
\midrule
Base Models & DeepSeek-R1-Distill-Qwen-7B, Qwen3-4B, Qwen3-8B, Llama-3.1-8B-Instruct & Learning Rate & $1 \times 10^{-5}$ \\ \addlinespace
Quantization & NF4 (4-bit) & LR Schedule & Cosine \\ \addlinespace
LoRA Rank ($r$) & 64 & Warmup Steps & 2 \\ \addlinespace
LoRA Alpha ($\alpha$) & 64 & Weight Decay & 0.1 \\ \addlinespace
GRPO Group Size ($G$) & 4 & Optimizer & AdamW (Torch) \\ \addlinespace
Max Sequence Length & 4096 & Adam $\beta_1, \beta_2$ & 0.9, 0.99 \\ \addlinespace
Max Prompt Length & 2048 & Max Completion Length & 2048 \\ \addlinespace
Per-device Batch Size & 4 & Gradient Accumulation & 1 \\ \addlinespace
Rollout Temperature & 0.7 & KL Penalty ($\beta$) & 0.01 \\ \addlinespace
Training Epochs & 5 & Clip Epsilon ($\varepsilon$) & 0.2 \\ \addlinespace
Max Grad Norm & 0.1 & Evaluation / Save Interval & 128 / 128 steps \\
\bottomrule
\end{tabularx}
\caption{Training and generation hyperparameters across model architectures.}
\label{tab:hyperparams}
\end{table*}

\subsection{Compute and Reproducibility}

All local training and evaluation were conducted on one NVIDIA GeForce RTX~5090 GPU with 32~GB GDDR7; no multi-GPU parallelism was used. The \texttt{gpt-oss-120b} reward judge was accessed remotely. We fix the pipeline-level random state to 3407 and the PyTorch and NumPy seeds to 42.

\FloatBarrier

\section{Reward Function Formulation}
\label{sec:appendix_rewardfn}

\subsection{Composite Objective}

For prompt $x$, completion $y$, ground truth $g$, and dataset $d$, let $\beta(y)$ and $\alpha(y)$ denote the text extracted from the \texttt{<think>} and \texttt{<answer>} spans. The reward passed to GRPO is
\begin{equation}
R=\frac{r_{\mathrm{cor}}+r_{\mathrm{cov}}+r_{\mathrm{dir}}}{3}.
\end{equation}
No additional transformation is applied before GRPO. Cor-GRPO uses only $r_{\mathrm{cor}}$.

\subsection{Correctness Reward}

\begin{equation}
r_{\mathrm{cor}}(x,y,g,d)=\mathbb{I}\!\left[V_d(\alpha(y),g)=1\right].
\end{equation}
If the \texttt{<answer>} span is absent, $r_{\mathrm{cor}}=0$. Table~\ref{tab:reward_verifiers} summarizes the dataset-specific verifiers.

\begin{table*}[t]
\centering
\scriptsize
\begin{tabularx}{\textwidth}{l X}
\toprule
\textbf{Dataset} & \textbf{Verifier} \\
\midrule
BalancedCOPA & Parse an integer and compare it with the one-based gold label. \\
CauseLogics & Uppercase prediction and gold label and require exact equality. \\
ClimateFever & Uppercase prediction and gold label and require exact equality among the four labels. \\
AbductionRules & Collapse whitespace, strip, lowercase, and require exact equality. \\
UniADILR & Extract standalone integers and compare their set with the gold proof-antecedent sentence identifiers; order and duplicates are ignored. \\
Crypto & Execute \texttt{transform(s)} on every held-out test input and require exact string outputs for all tests. \\
ListFunction & Execute \texttt{transform(lst)} on ten held-out test pairs and require matching integer-list outputs for all tests. \\
\bottomrule
\end{tabularx}
\caption{Dataset-specific correctness verifiers used during training.}
\label{tab:reward_verifiers}
\end{table*}

For executable tasks, code is extracted from the answer span, optional Markdown fences are removed, and each test runs in a separate process with a five-second timeout.

\subsection{Process Rewards}

The coverage judge receives $u$, $\beta(y)$, and a dataset-specific note and returns $m$ detail records with addressed indicators $z_j$. The implementation computes
\begin{equation}
r_{\mathrm{cov}}(x,y,d)=
\begin{cases}
\frac{1}{m}\sum_{j=1}^{m} z_j, & m>0,\\
0, & \text{otherwise}.
\end{cases}
\end{equation}
The denominator is the number of details returned by the judge; the code applies no additional alias merging or deduplication.

The directionality judge receives the same inputs and returns
\begin{equation}
r_{\mathrm{dir}}(x,y,d)\in\{0,0.5,1\},
\end{equation}
for explanation-to-evidence, mixed, and evidence-to-explanation reasoning, respectively. Invalid or unparseable judge outputs receive zero. If the \texttt{<think>} span is absent, the full completion is passed to the process judges. Both judges use \texttt{openai/gpt-oss-120b} at temperature $0.0$, and both process rewards are active for all seven training datasets.

\subsection{Training-Time Judge Prompts}
\label{app:training_reward_prompts}

The following prompts are used online to compute the process rewards. They are distinct from the generator prompts in Appendix~\ref{app:prompts} and the post-hoc evaluation prompts in Appendix~\ref{app:judge_prompts}.

\subsubsection{Evidence Coverage}

\paragraph{System message.}
\begin{Verbatim}[fontsize=\scriptsize,breaklines=true,breakanywhere=true]
## Training System Context

You are an automated reward judge embedded in a Group Relative Policy Optimization (GRPO) reinforcement-learning training loop. A language model (the *generator*) is being fine-tuned on abductive reasoning datasets. At each training step the generator produces a reasoning trace in response to an abductive problem; your structured score is used directly as a reward signal to update the generator's weights. Evaluate precisely and strictly from what is written in the trace.

The generator is learning abductive reasoning: given observations, identify the most plausible hypothesis and reason transparently from evidence to conclusion.

---

You are an expert evaluator of abductive reasoning traces.

## Your Task

Given a reasoning_trace in which a model selects one hypothesis to explain an observation, you must:

1. **Extract every specific detail** that appears in the observation (or the model's description of the observation) -- not just the main event, but also peripheral facts, contextual clues, timing details, quantities, locations, named entities, and any other particulars mentioned.

2. **For each detail**, decide whether the reasoning_trace *explicitly* connects that detail to the chosen hypothesis:
   - A detail is "addressed" only if there is a direct explanation of *why or how* the hypothesis connects to that detail.
   - Set addressed = false, evidence = an empty string if the detail is not explicitly connected.

3. **Write a brief overall_analysis** summarising the observation coverage.

## Important rules
- Be exhaustive across main and peripheral details.

## Output format

You may first think freely — identify each detail, deliberate on whether the reasoning trace addresses it, and work through any ambiguous cases. Once you have finished your analysis, place a single valid JSON block at the very end of your response with this structure:
{
  "overall_analysis": "Brief analysis of observation coverage in this reasoning_trace",
  "observation_details": [
    {
      "detail": "Specific atomic detail from the observation",
      "addressed": true,
      "evidence": "Quote from reasoning_trace showing how this detail is explained"
    }
  ]
}
\end{Verbatim}

\paragraph{User-message template.}
\begin{Verbatim}[fontsize=\scriptsize,breaklines=true,breakanywhere=true]
Dataset: {dataset_name}
Dataset context: {coverage_dataset_note}
Analyse the following observation and reasoning_trace and produce the structured observation-coverage evaluation.
<observation>
{policy_user_message}
</observation>
<reasoning_trace>
{extracted_think_span}
</reasoning_trace>
\end{Verbatim}

\subsubsection{Evidence--Explanation Directionality}

\paragraph{System message.}
\begin{Verbatim}[fontsize=\scriptsize,breaklines=true,breakanywhere=true]
## Training System Context

You are an automated reward judge embedded in a Group Relative Policy Optimization (GRPO) reinforcement-learning training loop. A language model (the *generator*) is being fine-tuned on abductive reasoning datasets. At each training step the generator produces a reasoning trace; your score is used directly as a reward signal to update the generator's weights.

The generator is learning that abductive reasoning runs strictly from evidence to explanation — not the reverse. Your score tells the training loop how well the current trace respects that directional constraint.

---

You are an expert evaluator assessing the logical directionality of reasoning chains generated by AI models.

## Metric Definition

**Evidence-Explanation Directionality:** The model demonstrates awareness that abduction runs strictly from evidence to explanation (not explanation to evidence / not deduction).

In proper abductive reasoning, the model MUST start from the known observations (the evidence) and logically reason forward to determine the best hypothesis (the explanation). Assuming an explanation is true and then checking whether it predicts or aligns with the given observations is a core logical flaw for this metric.

## Scoring Rubric

**Score 1.0 – Proper Abductive Direction**
- The model reasons from observations/evidence as the foundation.
- It builds its argument forward from the observations toward the hypothesis.
- Language cue: "Given these observations..."

**Score 0.5 – Ambiguous or Mixed Directionality**
- The model shows partial abduction but also some backwards reasoning.
- The logical flow is not clearly unidirectional.

**Score 0.0 – Explanation-Evidence Direction (deduction)**
- The model presupposes an explanation and checks if it predicts the observations.
- Reasons backwards from assumed conclusions to evidence.

## Output Format

You may first think freely — examine directional cues in the language, deliberate on which score best fits, and explain your reasoning. Once you have finished, place a single valid JSON block at the very end of your response:
{
  "directionality_score": 0.0,
  "explanation": "Brief explanation of the logical direction observed"
}

Note: directionality_score must strictly be 0.0, 0.5, or 1.0. Do not wrap the JSON in markdown code blocks.
\end{Verbatim}

\paragraph{User-message template.}
\begin{Verbatim}[fontsize=\scriptsize,breaklines=true,breakanywhere=true]
Dataset: {dataset_name}
Dataset context: {directionality_dataset_note}
Evaluate the reasoning chain below for Evidence-Explanation Directionality and assign a score of 0.0, 0.5, or 1.0.
**Observations / Evidence:**
<observations>
{policy_user_message}
</observations>
**Model's Reasoning Chain:**
<reasoning_chain>
{extracted_think_span}
</reasoning_chain>
Assign a directionality score (0.0 / 0.5 / 1.0) and briefly explain your assessment in the `explanation` field.
\end{Verbatim}

\subsubsection{Dataset-Specific Notes}

The strings below replace \texttt{coverage\_dataset\_note} and \texttt{directionality\_dataset\_note} in the templates above.

\paragraph{UniADILR.}
\textbf{Coverage:} The text passages and numbered sentences are the observation. Extract all factual claims together with their sentence identifiers.
\textbf{Directionality:} The numbered Context sentences are the evidence, and the stated Hypothesis is the candidate explanation. Assess whether the trace reasons from the context evidence toward support for that hypothesis.

\paragraph{BalancedCOPA.}
\textbf{Coverage:} The 'Premise:' field is the observation. Extract all stated facts and contextual details from the premise.
\textbf{Directionality:} The Premise is the observed event, and Choice 1 and Choice 2 are candidate causes. Assess whether the trace reasons from the premise toward selecting the more plausible cause.

\paragraph{CauseLogics.}
\textbf{Coverage:} The Premises, Rules, and Phenomenon form the observation, while the Possible Cause is the candidate hypothesis. Extract all stated facts, rule conditions and conclusions, and phenomenon details.
\textbf{Directionality:} The Premises and Rules are background evidence, the Phenomenon is the observation, and the Possible Cause is the candidate explanation. Assess whether the trace reasons from the stated evidence toward evaluating that explanation.

\paragraph{ClimateFever.}
\textbf{Coverage:} The Claim and provided Evidence sentences together form the observation. Extract all factual statements and their individual components.
\textbf{Directionality:} Treat the provided Evidence sentences as the observations and the selected claim label as the conclusion. Assess whether the reasoning moves from the evidence toward that label.

\paragraph{AbductionRules.}
\textbf{Coverage:} The Context facts and rules together with the Query form the observation. Extract all relevant conditions, entity properties, stated facts, and query details.
\textbf{Directionality:} Treat the Context facts, rules, and Query as the observations and the proposed missing fact as the hypothesis. Assess whether the reasoning moves from the given context and query toward that fact.

\paragraph{ListFunction.}
\textbf{Coverage:} The ten training input-output pairs are the observation. Treat each demonstrated list transformation as a detail and assess whether the reasoning explains the pattern across all examples.
\textbf{Directionality:} The observed list input-output pairs are the evidence, and the inferred list transformation is the explanation. Assess whether the reasoning moves from the examples to the rule.

\paragraph{Crypto.}
\textbf{Coverage:} The ten training input-output pairs are the observation. Treat each demonstrated string transformation as a detail and assess whether the reasoning explains the character-level pattern across all examples.
\textbf{Directionality:} The observed string input-output pairs are the evidence, and the inferred string transformation is the explanation. Assess whether the reasoning moves from the examples to the rule.

\subsection{Reward Metric Selection}
\label{app:reward_selection}

To select process-level metrics for reward shaping, we screened each candidate metric along two axes: task applicability and reward-hacking risk. Applicability measures whether improving a metric is likely to support the reasoning behavior required by each training source. Reward-hacking risk measures how easily a policy could increase the metric through superficial changes in trace style, verbosity, or formatting rather than through genuine improvements in reasoning.

Table~\ref{tab:reward_metric_screen} summarizes this screening process. Dark green cells indicate that a metric has a direct and consistent effect on the corresponding dataset, while hatched light green cells indicate a moderate or context-dependent effect. White cells indicate that the metric is negligible or not applicable for that dataset.

This screening led us to select Evidence Coverage and Evidence--Explanation Directionality as our primary process rewards. Evidence Coverage is broadly applicable because it encourages the model to explicitly account for the observations provided in the prompt across selection, verification, and generation settings. Evidence--Explanation Directionality complements this by encouraging the reasoning trace to preserve the abductive flow from evidence toward explanation. In contrast, although metrics such as \textit{Backtracking} and \textit{Uncertainty Markers} can be useful for analysis, they carry substantially higher reward-hacking risk: a policy could increase them by producing artificial self-corrections or excessive hedging without improving the underlying reasoning process.

\definecolor{metricStrong}{HTML}{00B050}
\definecolor{metricSome}{HTML}{C6E0B4}
\definecolor{riskVeryHigh}{HTML}{C00000}
\definecolor{riskHigh}{HTML}{FF0000}
\definecolor{riskModerate}{HTML}{FFC000}
\definecolor{riskLowMed}{HTML}{FFFF00}
\definecolor{riskVeryLow}{HTML}{92D050}

\newlength{\metricColW}
\setlength{\metricColW}{4.05cm}
\newlength{\rewardCellW}
\setlength{\rewardCellW}{1.40cm}
\newlength{\riskColW}
\setlength{\riskColW}{1.80cm}
\newlength{\rewardRowH}
\setlength{\rewardRowH}{5.0ex}
\newlength{\rewardHeadH}
\setlength{\rewardHeadH}{5.0ex}

\newcommand{\MetricName}[1]{%
    \parbox[c][\rewardRowH][c]{\metricColW}{\raggedright #1}%
}
\newcommand{\DataHead}[1]{%
    \parbox[c][\rewardHeadH][c]{\rewardCellW}{\centering\scriptsize\bfseries #1}%
}
\newcommand{\RiskHead}{%
    \parbox[c][\rewardHeadH][c]{\riskColW}{\centering\scriptsize\bfseries Reward-hacking\\risk}%
}

\newcommand{\Eff}{%
    \begin{tikzpicture}[baseline=(cell.center)]
        \node[
            inner sep=0pt,
            outer sep=0pt,
            minimum width=\rewardCellW,
            minimum height=\rewardRowH,
            fill=metricStrong
        ] (cell) {};
    \end{tikzpicture}%
}

\newcommand{\Some}{%
    \begin{tikzpicture}[baseline=(cell.center)]
        \node[
            inner sep=0pt,
            outer sep=0pt,
            minimum width=\rewardCellW,
            minimum height=\rewardRowH
        ] (cell) {};
        \fill[metricSome] (cell.south west) rectangle (cell.north east);
        \fill[
            pattern=north east lines,
            pattern color=black!45
        ] (cell.south west) rectangle (cell.north east);
    \end{tikzpicture}%
}

\newcommand{\None}{%
    \begin{tikzpicture}[baseline=(cell.center)]
        \node[
            inner sep=0pt,
            outer sep=0pt,
            minimum width=\rewardCellW,
            minimum height=\rewardRowH
        ] (cell) {};
    \end{tikzpicture}%
}

\newcommand{\RiskCell}[3]{%
    \begin{tikzpicture}[baseline=(cell.center)]
        \node[
            inner sep=0pt,
            outer sep=0pt,
            minimum width=\riskColW,
            minimum height=\rewardRowH,
            text width=\riskColW,
            align=center,
            fill=#1,
            text=#2,
            font=\bfseries
        ] (cell) {#3};
    \end{tikzpicture}%
}

\newcommand{\RiskVH}{\RiskCell{riskVeryHigh}{white}{V. high}}
\newcommand{\RiskH}{\RiskCell{riskHigh}{black}{High}}
\newcommand{\RiskM}{\RiskCell{riskModerate}{black}{Mod.}}
\newcommand{\RiskLM}{\RiskCell{riskLowMed}{black}{Low--med.}}
\newcommand{\RiskVL}{\RiskCell{riskVeryLow}{black}{V. low}}

\begin{table*}[h!]
\centering
\small
\setlength{\tabcolsep}{0pt}
\renewcommand{\arraystretch}{1.0}

\begin{tabular}{
@{}
>{\raggedright\arraybackslash}m{\metricColW}
*{7}{>{\centering\arraybackslash}m{\rewardCellW}}
>{\centering\arraybackslash}m{\riskColW}
@{}
}
\toprule
\parbox[c][\rewardHeadH][c]{\metricColW}{\bfseries Metric}
&
\DataHead{UniADILR}
&
\DataHead{COPA}
&
\DataHead{Cause\\Logics}
&
\DataHead{CLIMATE}
&
\DataHead{Abduction\\Rules}
&
\DataHead{List\\Func.}
&
\DataHead{Crypto}
&
\RiskHead
\\
\midrule

\MetricName{Branchiness}
& \Eff & \Eff & \None & \Some & \Eff & \Some & \Some & \RiskH \\

\MetricName{Backtracking}
& \Some & \Some & \None & \Some & \Eff & \Eff & \Eff & \RiskVH \\

\MetricName{Differential Elimination}
& \Eff & \Eff & \Some & \Some & \Some & \Some & \Some & \RiskM \\

\MetricName{Prior Invocation}
& \Some & \Eff & \None & \Some & \None & \None & \Some & \RiskH \\

\MetricName{Evidence Coverage}
& \Eff & \Eff & \Eff & \Eff & \Eff & \Eff & \Eff & \RiskLM \\

\MetricName{Evidence--Explanation Directionality}
& \Eff & \Some & \Some & \Eff & \Eff & \Eff & \Eff & \RiskLM \\

\MetricName{Uncertainty Markers}
& \Eff & \Eff & \None & \Some & \Some & \Some & \Some & \RiskVH \\

\bottomrule
\end{tabular}

\vspace{3pt}
\begin{minipage}{0.97\textwidth}
\footnotesize
\textbf{Legend.}
\tikz[baseline=-0.45ex]{
    \fill[metricStrong] (0,0) rectangle (1.35em,0.8em);
}
= effective candidate signal;
\tikz[baseline=-0.45ex]{
    \fill[metricSome] (0,0) rectangle (1.35em,0.8em);
    \fill[pattern=north east lines, pattern color=black!45]
        (0,0) rectangle (1.35em,0.8em);
}
= somewhat effective or context-dependent signal;
\tikz[baseline=-0.45ex]{
    \draw[black!35] (0,0) rectangle (1.35em,0.8em);
}
= not applicable or negligible. Reward-hacking risk estimates how easily a metric could be maximized by changing the style or verbosity of the trace rather than improving the underlying reasoning behavior.
\end{minipage}

\caption{
Screening candidate process-level metrics as possible reward terms. Applicability is evaluated per training source, while reward-hacking risk is assessed at the metric level. This screening motivates our reward-composition choice: Evidence Coverage and Evidence--Explanation Directionality are selected because they are broadly applicable across datasets and comparatively resistant to superficial optimization.
}
\label{tab:reward_metric_screen}
\end{table*}

\section{Process-Level Metrics}
\label{app:process_level_metrics}
\newcommand{\tabrot}[1]{\rotatebox{90}{\textbf{#1}}}

\subsection{Evaluation Framework}
\label{app:eval_framework}

All seven process-level metrics are evaluated using an LLM-as-judge approach.
Each reasoning trace is passed to \textbf{Gemini~3~Flash} together with a
structured system prompt specific to the metric and a short dataset-specific
contextual framing.  The judge is required to return a fixed JSON schema;
output parsing is fully deterministic and involves no free-text post-processing.
No few-shot demonstrations are used in any judge prompt. (Note that this offline evaluation setup uses Gemini~3~Flash to independently verify trace quality, which is distinct from the training phase where \texttt{gpt-oss-120b} was used to compute dynamic RL rewards).

Three families of metric are used.
\textbf{Counting metrics} (Uncertainty Markers, Branchiness, Backtracking,
Differential Elimination, Prior Invocation) instruct the judge to return a list
of concrete $\langle \textit{excerpt}, \textit{explanation} \rangle$ pairs, one per identified
occurrence.  The reported value per trace is the total number 
of extracted instances, averaged over all traces in a dataset.
\textbf{Coverage metrics} (Evidence Coverage) ask the judge to enumerate
every atomic observation detail and mark each one as explicitly addressed or
not; the reported score per trace is the proportion of total details that are 
explicitly addressed, averaged over the dataset.
\textbf{Score-based metrics} (Evidence--Explanation Directionality) ask the
judge to assign one of three discrete values $(0.0,\ 0.5,\ 1.0)$; the
reported value is the per-trace score averaged over the dataset.

\subsection{Metric Definitions}
\label{app:metric_defs}

\subsubsection{Uncertainty Markers}
\label{app:metric_uncertainty}

\noindent\textbf{Type:} Counting.

\noindent\textbf{Core question:} \textit{How many individual probabilistic or hedging
words and phrases appear in the reasoning trace?}

Uncertainty Markers extracts every distinct occurrence of a word or phrase
that signals epistemic hedging or probabilistic qualification, rather than
asking a binary question about their presence.  Each occurrence is logged as
an independent example.

\noindent\textbf{Counted:}
\begin{itemize}[noitemsep, topsep=3pt, leftmargin=*]
  \item Probability/likelihood qualifiers: \textit{probably, likely, unlikely, possibly, in all likelihood, with high probability, most likely}
  \item Epistemic modal verbs: \textit{might, may, could, seems to, appears to, tends to}
  \item Hedging phrases: \textit{I believe, I think, I suspect, this suggests, this may indicate}
  \item Degree/approximation qualifiers: \textit{approximately, roughly, to some extent, somewhat, fairly}
  \item Epistemic uncertainty statements: \textit{we cannot be sure, it is unclear, the evidence is inconclusive, this remains to be confirmed}
  \item Frequency and scope limiters: \textit{often, typically, generally, sometimes, in some cases, less common}
\end{itemize}

\noindent\textbf{Not Counted:}
\begin{itemize}[noitemsep, topsep=3pt, leftmargin=*]
  \item Objective risk or statistical labels (\emph{e.g.}\ ``high risk'')
  \item Phrases that establish a premise boundary (\emph{e.g.}\ ``based on the information provided'')
  \item Evaluative or affective state descriptions (\emph{e.g.}\ ``there is concern'')
\end{itemize}

\noindent\textbf{Reported statistic:} Mean count of marker occurrences per
trace, averaged over the dataset.

\subsubsection{Branchiness}
\label{app:metric_branchiness}

\noindent\textbf{Type:} Counting.

\noindent\textbf{Core question:} \textit{How many times does the reasoning genuinely
explore multiple distinct candidate explanations for the same observation?}

Branchiness measures whether the model considers substantively different
explanatory candidates rather than following a single linear chain.  The
competing candidates must differ in underlying causal mechanism, agent,
domain, or interpretation --- not merely in phrasing or confidence level.

\noindent\textbf{Counted:}
\begin{itemize}[noitemsep, topsep=3pt, leftmargin=*]
  \item Exploring two diagnoses with separate evidential implications (\textit{``If condition X we would expect F; if condition Y we would expect G''})
  \item Identifying different causal mechanisms for the same narrative event
  \item Building and comparing competing hypotheses
\end{itemize}

\noindent\textbf{Not counted:}
\begin{itemize}[noitemsep, topsep=3pt, leftmargin=*]
  \item Refinements or phrasings of the same explanation
  \item Forward-branching planning logic
  \item Procedural trial-and-error
  \item The final answer selection
  \item A brief mention of an alternative immediately rejected
  \item Simple step-by-step narration
  \item Restating the supplied answer options
\end{itemize}

\noindent\textbf{Reported statistic:} Mean count of branching instances per
trace, averaged over the dataset.

\subsubsection{Backtracking}
\label{app:metric_backtracking}

\noindent\textbf{Type:} Counting.

\noindent\textbf{Core question:} \textit{How many times does the reasoning explicitly
identify an error or flaw and change direction?}

Backtracking (self-correction) captures deliberate revision: the model
recognises that something it said or computed is wrong and reverses course.
This is distinct from Branchiness, which measures exploration among valid
alternatives.

\noindent\textbf{Counted:}
\begin{itemize}[noitemsep, topsep=3pt, leftmargin=*]
  \item Explicit admission of error (\textit{``Wait, that's wrong'', ``I made a mistake''})
  \item Deliberate pausing and restarting (\textit{``Hold on, let me re-read the problem''})
  \item Mid-reasoning strategy change (\textit{``On second thought\ldots'', ``That approach doesn't work, so\ldots''})
  \item Realisation of a missed detail (\textit{``I realise I forgot to account for\ldots''})
\end{itemize}

\noindent\textbf{Not counted:}
\begin{itemize}[noitemsep, topsep=3pt, leftmargin=*]
  \item Comparing two valid paths (Branchiness)
  \item A ``however'' contrast without admitting an error
  \item The final answer selection
\end{itemize}

\noindent\textbf{Reported statistic:} Mean count of backtracking instances per
trace, averaged over the dataset.

\subsubsection{Differential Elimination}
\label{app:metric_diff_elim}

\noindent\textbf{Type:} Counting.

\noindent\textbf{Core question:} \textit{How many explicit elimination or refutation
moves against alternatives appear in the reasoning trace?}

Differential Elimination extracts each distinct case where the model rules out
an alternative hypothesis, answer option, or interpretation with an explicit
evidence-grounded reason.  The refutation must supply a \emph{why}, not merely
assert that an option is incorrect.

\noindent\textbf{Counted:}
\begin{itemize}[noitemsep, topsep=3pt, leftmargin=*]
  \item \textit{``We can rule out A because it contradicts symptom X''}
  \item Conditional falsification (\textit{``If B were true we would see Y, but we do not''})
  \item Explicit contradiction between an alternative and the observed details
\end{itemize}

\noindent\textbf{Not counted:}
\begin{itemize}[noitemsep, topsep=3pt, leftmargin=*]
  \item Empty dismissals without a context-grounded reason
  \item Pure positive support for the chosen hypothesis
  \item Listing options without evaluating them
  \item Vague preference statements
\end{itemize}

\noindent\textbf{Reported statistic:} Mean count of elimination instances per
trace, averaged over the dataset.

\subsubsection{Prior Invocation}
\label{app:metric_prior}

\noindent\textbf{Type:} Counting.

\noindent\textbf{Core question:} \textit{How many times does the reasoning explicitly
invoke prior probability, typicality, or base-rate knowledge?}

Prior Invocation captures explicit references to background frequency or
likelihood that the model brings to bear on the problem before or alongside
the specific evidence in the trace.

\noindent\textbf{Counted:}
\begin{itemize}[noitemsep, topsep=3pt, leftmargin=*]
  \item Population base rates
  \item Domain-knowledge priors (\textit{``Most patients with these symptoms have condition X''})
  \item Comparative likelihoods (\textit{``X is far more common than Y''})
  \item General tendencies or patterns (\textit{``Usually, this symptom indicates\ldots''})
  \item Explicit Bayesian-style reasoning that weighs a prior against new evidence
\end{itemize}

\noindent\textbf{Not counted:}
\begin{itemize}[noitemsep, topsep=3pt, leftmargin=*]
  \item Conclusions drawn only from the specific observations in the problem (posterior inferences)
  \item World-knowledge facts without frequency or likelihood content
  \item The model's own hedging expressions, which are captured by Uncertainty Markers
  \item Restatements of information given in the prompt
\end{itemize}

\noindent\textbf{Reported statistic:} Mean count of prior-invocation instances
per trace, averaged over the dataset.

\subsubsection{Evidence Coverage}
\label{app:metric_coverage}

\noindent\textbf{Type:} Coverage.

\noindent\textbf{Core question:} \textit{What fraction of the specific observation
details are explicitly accounted for by the chosen hypothesis?}

Evidence Coverage enumerates every atomic observation detail present in the
input and assesses whether the reasoning trace explicitly connects each one to
the chosen hypothesis.  Atomicity is enforced throughout: compound statements
are split into indivisible facts (\emph{e.g.}\ ``headache and vomiting''
becomes two items; ``Kernig and Brudzinski signs present'' becomes two items;
every medication, vital sign, and lab value is its own item).  A detail is
marked \emph{addressed} only when the trace makes a clear logical link between
that detail and the chosen hypothesis --- restating the detail or merely
acknowledging its existence is not sufficient.

The coverage score per trace is
\[
  s_{\mathrm{cov}} \;=\;
  \frac{\#\text{addressed details}}{\#\text{total details}} \;\in [0,1].
\]

\noindent\textbf{Reported statistic:} Mean $s_{\mathrm{cov}}$ over all traces
in the dataset, expressed as a percentage in tables.

\subsubsection{Evidence--Explanation Directionality}
\label{app:metric_directionality}

\noindent\textbf{Type:} Score-based.

\noindent\textbf{Core question:} \textit{How strongly does the reasoning respect the
abductive direction from evidence to explanation?}

This metric assesses whether the model treats the given observations as the
fixed starting point and reasons forward to the best explanatory hypothesis.
The core failure mode the metric is designed to detect is the logical reversal
in which a model assumes a hypothesis first and then deductively checks whether
it predicts the evidence --- a pattern corresponding to the prediction format
``If hypothesis $H$ is true, we would expect observation $O$; we observe $O$,
therefore $H$.''

The judge assigns exactly one of three discrete scores:
\begin{itemize}[nosep]
  \item \textbf{1.0} --- Clear evidence\,$\to$\,explanation reasoning: the
    trace treats the observations as given, asks what best explains them, and
    builds its argument from the evidence upward to the conclusion.
  \item \textbf{0.5} --- Ambiguous or mixed directionality: the reasoning
    connects evidence and explanation but the logical flow is unclear or
    loosely structured, wavering between both directions.
  \item \textbf{0.0} --- Backward or deductive direction: the trace assumes a
    conclusion first and then verifies that the evidence confirms it.
\end{itemize}

\noindent\textbf{Reported statistic:} Mean directionality score over all
traces in the dataset (range 0.0--1.0).

\subsection{Dataset-Specific Judge Notes}
\label{app:dataset_notes}

Every judge prompt injects a short dataset-specific note providing contextual framing 
for the metric definition based on the format and content of each evaluation split.
Table~\ref{tab:notes_all_metrics} gives the notes for all seven metrics (the five counting metrics, 
Evidence Coverage, and Evidence--Explanation Directionality).

\subsection{Judge Prompts}
\label{app:judge_prompts}

The subsections below give the verbatim system prompt and user-prompt
template sent to Gemini~3~Flash for each metric.  The placeholder
\texttt{\{dataset\_specific\_note\}} is replaced at runtime with the
corresponding cell from Table~\ref{tab:notes_all_metrics}.  All evaluations 
are strictly zero-shot.

\subsubsection{Uncertainty Markers}
\label{app:prompt_uncertainty}

\noindent\textbf{System Prompt:}
\begin{lstlisting}[breaklines=true, basicstyle=\small\ttfamily]
You are an expert linguistic analyst evaluating AI-generated reasoning traces.

## What is an Uncertainty Marker?

An uncertainty marker is a **specific word or phrase** that signals the model is
expressing a degree of belief, possibility, probability, rather than stating
something as an absolute, universal, or established fact. Your job is to locate
every individual marker that appears in the reasoning trace.

## Categories of uncertainty markers to extract

### 1. Probability / likelihood qualifiers
Words or phrases that place something on a probability scale.
Examples: "probably", "likely", "unlikely", "possibly", "conceivably",
"in all likelihood", "there is a chance", "with high probability",
"most likely", "least likely", "more probable than".

### 2. Epistemic modals and verbs of potential
Verbs that express possibility, tentative judgement, or potential rather than
a guaranteed outcome.
Examples: "might", "may", "could", "would", "can" (when used as 'has the
potential to', e.g., "can help"), "seems to", "appears to", "tends to".

### 3. Hedging phrases (first-person or impersonal)
Phrases that explicitly frame a statement as a belief or estimate.
Examples: "I believe", "I think", "I suspect", "I'm not certain but",
"it is possible that", "it seems that", "it appears that",
"this suggests", "this may indicate".

### 4. Degree / approximation qualifiers
Phrases that soften a claim by expressing partial knowledge or approximation.
Examples: "approximately", "roughly", "around", "about", "or so",
"to some extent", "in part", "somewhat", "fairly", "relatively".

### 5. Epistemic uncertainty statements
Explicit acknowledgements that something is unknown or unconfirmed.
Examples: "we cannot be sure", "it is uncertain whether", "it is unclear",
"the evidence is inconclusive", "this is not definitively established",
"this remains to be confirmed".

### 6. Frequency and scope limiters
Words that soften a universal assertion by limiting its frequency or scope,
leaving room for exceptions.
Examples: "often", "typically", "generally", "frequently", "less common",
"in some cases", "sometimes", "usually".

## What NOT to Extract (False Positives)

Do **NOT** extract the following linguistic constructs, as they do not represent
epistemic uncertainty:

- **Objective risk or statistical metrics:** Mentions of "risk" describe an
  objective state or classification, not the speaker's doubt.
  (e.g., Do NOT extract "high risk", "reduces the risk").
- **Evidential attributions / Premise boundaries:** Phrases that cite a source
  or establish the boundary of the premise.
  (e.g., Do NOT extract "Based on the information provided",
  "According to the text").
- **Evaluative or affective states:** Stating that an emotion or clinical
  attitude exists is a factual claim about a state of affairs.
  (e.g., Do NOT extract "There is concern", "It is alarming").

## Extraction rules

- Extract **each individual marker occurrence** as a separate example, even if
  the same word appears multiple times. Every occurrence is its own entry.
- The `excerpt` must be a **short, direct quote** from the text -- ideally the
  single word or short phrase itself, plus just enough surrounding context
  (<= 15 words) to make it readable.
- The `explanation` must name the marker category (from the list above) and
  briefly state what belief, probability, or limitation the marker expresses
  in context.
- If the same sentence contains two distinct markers, extract them as two
  separate entries.
- Do **not** paraphrase or alter the quoted text.

## Dataset-specific note (current dataset only)

{dataset_specific_note}

## JSON output format

Return ONLY valid JSON with this structure:
{
  "overall_analysis": "Brief analysis of uncertainty markers density in this
    reasoning trace",
  "examples": [
    {
      "excerpt": "Quote of the uncertainty marker from the reasoning trace",
      "explanation": "Category and meaning of this uncertainty marker"
    }
  ]
}
\end{lstlisting}

\vspace{0.8em}
\noindent\textbf{User Prompt Template:}
\begin{lstlisting}[breaklines=true, basicstyle=\small\ttfamily]
Dataset: {dataset}

Extract every individual uncertainty marker from the following reasoning trace.
Return one entry per marker occurrence.

<reasoning_trace>
{text}
</reasoning_trace>
\end{lstlisting}

\subsubsection{Branchiness}
\label{app:prompt_branchiness}

\noindent\textbf{System Prompt:}
\begin{lstlisting}[breaklines=true, basicstyle=\small\ttfamily]
You are an expert reasoning analyst evaluating AI-generated reasoning traces.

## What is Branchiness?

Branchiness measures whether the reasoning **genuinely explores multiple
distinct candidate explanations** for the same observation before settling on
one, rather than following a single linear path.

The key distinction is this:
- Count multiple candidate explanations only when they are substantively
  different explanations of the observation.
- Do NOT count multiple versions, refinements, or restatements of the same
  underlying explanation.

## What COUNTS as a branching moment

Extract an example when you see:
1. Exploring two or more genuinely distinct candidate explanations for the same
   observation before settling on one.
2. Identifying different causal mechanisms, agents, domains, or scenario
   interpretations that could explain the observation.
3. Building and comparing competing hypotheses with their implications/evidence
   ("If diagnosis X we'd expect F... If diagnosis Y we'd expect G...").

## What does NOT count

- Multiple phrasings, refinements, or confidence adjustments of the same
  explanation.
- A main explanation plus a small modifier or detail added to that same
  explanation.
- Strictly forward-branching predictive logic or conditional planning
  (e.g., "If I do X, then Y happens").
- Trying different procedural solution methods (this is not abductive
  branching).
- The final answer selection or conclusion.
- A brief mention of an alternative followed by immediate rejection with no
  exploration.
- Simple step-by-step narration (First / Next / Then).
- Listing the given answer options without exploring them.

## Dataset-specific note (current dataset only)

{dataset_specific_note}

## Extraction rules

- Extract each distinct branching moment as a separate example.
- Use `excerpt` as a short direct quote from the reasoning trace
  (preferably <= 25 words).
- Use `explanation` to state why that quote reflects multiple genuinely
  distinct candidate explanations rather than variants of the same explanation
  or linear narration.
- If the same branch is repeated with no new reasoning content, extract it
  once.
- Do not count superficial variation unless the competing explanations differ
  in underlying mechanism, agent, domain, or interpretation.
- Do not paraphrase quoted text.

## JSON output format

Return ONLY valid JSON with this structure:
{
  "overall_analysis": "Brief analysis of branchiness in this reasoning trace",
  "examples": [
    {
      "excerpt": "Quote of the branching moment from the reasoning trace",
      "explanation": "Why this represents branching"
    }
  ]
}
\end{lstlisting}

\vspace{0.8em}
\noindent\textbf{User Prompt Template:}
\begin{lstlisting}[breaklines=true, basicstyle=\small\ttfamily]
Dataset: {dataset}

Analyze the following reasoning trace for Branchiness and extract concrete
examples.

<reasoning_trace>
{text}
</reasoning_trace>
\end{lstlisting}

\subsubsection{Backtracking}
\label{app:prompt_backtracking}

\noindent\textbf{System Prompt:}
\begin{lstlisting}[breaklines=true, basicstyle=\small\ttfamily]
You are an expert reasoning analyst evaluating AI-generated reasoning traces.

## What is Backtracking?

Backtracking (also called Self-Correction) occurs when the reasoning explicitly
updates or abandons a previously stated explanatory hypothesis in light of
newly processed evidence or a deeper logical mismatch, and then changes
direction.

## What COUNTS as a backtracking moment

Extract an example when you see:
- Explicit admission of error: "Wait, that's wrong", "I made a mistake",
  "Actually, I need to reconsider ..."
- Deliberate pausing and restarting: "Hold on, let me re-read the problem",
  "Let's go back to step 2"
- A change of strategy mid-reasoning: "Instead, let's try ...",
  "On second thought ...", "That approach doesn't work, so ..."
- Realisation of a missed detail: "I realise I forgot to account for ...",
  "This doesn't look right because ..."

## What does NOT count

- Comparing two valid paths (that is Branchiness).
- A simple "However" that introduces a contrast without admitting an error.
- The final answer selection.

## Dataset-specific note (current dataset only)

{dataset_specific_note}

## Extraction rules

- Extract each explicit self-correction/backtracking event as a separate
  example.
- Use `excerpt` as a short direct quote from the reasoning trace
  (preferably <= 25 words).
- Use `explanation` to clarify what was revised and why this is true
  backtracking.
- Do not count simple contrast words unless they indicate an actual
  correction.
- Do not paraphrase quoted text.

## JSON output format

Return ONLY valid JSON with this structure:
{
  "overall_analysis": "Brief analysis of backtracking/self-correction in
    this reasoning trace",
  "examples": [
    {
      "excerpt": "Quote of the backtracking moment from the reasoning trace",
      "explanation": "Why this represents backtracking/self-correction"
    }
  ]
}
\end{lstlisting}

\vspace{0.8em}
\noindent\textbf{User Prompt Template:}
\begin{lstlisting}[breaklines=true, basicstyle=\small\ttfamily]
Dataset: {dataset}

Analyze the following reasoning trace for Backtracking / Self-Correction and
extract concrete examples.

<reasoning_trace>
{text}
</reasoning_trace>
\end{lstlisting}

\subsubsection{Differential Elimination}
\label{app:prompt_diff_elim}

\noindent\textbf{System Prompt:}
\begin{lstlisting}[breaklines=true, basicstyle=\small\ttfamily]
You are an expert evaluator of abductive reasoning traces.

## What is Differential Elimination?

Differential Elimination measures how many distinct alternatives are explicitly
rejected or ruled out during reasoning. Unlike a binary presence/absence check,
this metric extracts each elimination instance as its own example.

## What COUNTS as a differential elimination instance

Extract an example when the trace explicitly:
1. Rules out an alternative hypothesis/option with a specific reason.
2. Shows contradiction between an alternative and observed details.
3. Uses conditional falsification
   ("If X were true, we would see Y, but we don't.").
4. Compares alternatives and explicitly marks one as less plausible or
   incompatible.

## What does NOT count

- Empty dismissals (e.g., simply stating "Option B is incorrect" or
  "Option C is false") without providing a context-grounded "why" based on
  the specific evidence.
- Pure support for the chosen hypothesis without discussing alternatives.
- Listing options without evaluating or eliminating them.
- Vague preference statements without a concrete elimination reason.
- Final answer statements that do not include explicit refutation content.

## Dataset-specific note (current dataset only)

{dataset_specific_note}

## Extraction rules

- Extract each distinct elimination/refutation event as a separate example.
- Use `excerpt` as a short direct quote from the reasoning trace
  (preferably <= 30 words).
- Use `explanation` to state what alternative was eliminated and why.
- If the same elimination is repeated without new rationale, include it once.
- Do not paraphrase quoted text.

## JSON output format

Return ONLY valid JSON with this structure:
{
  "overall_analysis": "Brief analysis of elimination behavior in this
    reasoning trace",
  "examples": [
    {
      "excerpt": "Quote showing explicit elimination of an alternative",
      "explanation": "What was eliminated and why this is a valid elimination
        instance"
    }
  ]
}
\end{lstlisting}

\vspace{0.8em}
\noindent\textbf{User Prompt Template:}
\begin{lstlisting}[breaklines=true, basicstyle=\small\ttfamily]
Dataset: {dataset}

Extract every explicit Differential Elimination instance from the following
reasoning trace. Return one entry per distinct elimination event.

<reasoning_trace>
{text}
</reasoning_trace>
\end{lstlisting}

\subsubsection{Prior Invocation}
\label{app:prompt_prior}

\noindent\textbf{System Prompt:}
\begin{lstlisting}[breaklines=true, basicstyle=\small\ttfamily]
You are an expert analyst evaluating AI-generated reasoning traces.

## What is a Prior?

A prior (or prior probability / base rate) is a pre-existing probability,
frequency, or background knowledge about how common or likely something is
in general, *before* considering the specific observations at hand. The
reasoner uses this prior to adjust their final judgment.

Your task is to identify every instance where the model explicitly brings
in such prior information to inform its reasoning.

## Categories of priors to extract

### 1. Population base rates
Statistical information about how common a condition, event, or outcome is
in a relevant population.
Examples: "This disease affects 1 in 10,000 people", "The prevalence of
this condition is approximately 5%", "This is a rare disorder", "This is
a common occurrence in this age group".

### 2. Prior probabilities from domain knowledge
General knowledge about likelihoods that the model brings to bear on the
problem, not derived from the specific observations.
Examples: "Most patients with these symptoms have condition X", "Typically,
this type of failure is caused by Y", "In general, Z is more likely than W".

### 3. Comparative likelihoods
Explicit comparisons of how likely different possibilities are, based on
background knowledge rather than the specific evidence.
Examples: "X is far more common than Y", "This explanation is more probable
a priori", "Without specific evidence, we would expect Z".

### 4. Reference to general tendencies or patterns
References to what "usually" or "typically" happens, used as a prior to
guide reasoning.
Examples: "Usually, this symptom indicates...", "Typically, patients with
this profile...", "In most cases like this...".

### 5. Explicit Bayesian-style reasoning
Cases where the model explicitly weighs prior probability against new evidence.
Examples: "Even though the test is positive, given the low base rate...",
"The prior probability is low, so we need strong evidence...", "Combining
the prior with these observations...".

## Extraction rules

- Extract **each distinct prior consideration** as a separate example.
- The `excerpt` must be a **short, direct quote** from the text that shows
  the model referencing prior information (<= 25 words of context).
- The `explanation` must identify the type of prior (from the categories
  above) and briefly explain what prior probability or base rate is being
  referenced.
- If the model mentions the same prior multiple times in different parts
  of the reasoning, extract each occurrence separately.
- Do **not** paraphrase or alter the quoted text.

## What does NOT count as a prior

- Conclusions drawn *only* from the specific observations in the problem
  (these are posterior inferences, not priors).
- General knowledge that doesn't involve probability or frequency (e.g.,
  "The heart pumps blood" is a fact, not a prior).
- Hypotheses generated during reasoning without reference to their general
  likelihood.
- The model's own uncertainty expressions (e.g., "I think", "probably") --
  those are captured by the uncertainty_markers metric.
- Restatements of information given in the problem prompt.

## Dataset-specific note (current dataset only)

{dataset_specific_note}

## JSON output format

Return ONLY valid JSON with this structure:
{
  "overall_analysis": "Brief analysis of prior probability usage in this
    reasoning trace",
  "examples": [
    {
      "excerpt": "Quote of the prior probability consideration from the
        reasoning trace",
      "explanation": "Type of prior and what probability/frequency is being
        referenced"
    }
  ]
}
\end{lstlisting}

\vspace{0.8em}
\noindent\textbf{User Prompt Template:}
\begin{lstlisting}[breaklines=true, basicstyle=\small\ttfamily]
Dataset: {dataset}

Extract every instance where the model considers a prior probability or
base rate in the following reasoning trace. Return one entry per prior
consideration.

<reasoning_trace>
{text}
</reasoning_trace>
\end{lstlisting}

\subsubsection{Evidence Coverage}
\label{app:prompt_coverage}

\noindent\textbf{System Prompt:}
\begin{lstlisting}[breaklines=true, basicstyle=\small\ttfamily]
You are an expert evaluator of abductive reasoning traces.

## Your Task

Given a reasoning_trace in which a model selects one hypothesis to explain an
observation, you must:

1. **Extract every specific detail** that appears in the observation (or the
   model's description of the observation) -- not just the main event, but
   also peripheral facts, contextual clues, timing details, quantities,
   locations, named entities, and any other particulars mentioned.

2. **For each detail**, decide whether the reasoning_trace *explicitly*
   connects that detail to the chosen hypothesis. A detail is "addressed" only
   if the reasoning_trace makes a clear logical link between that detail and
   the hypothesis -- not merely restating it or acknowledging it exists.

3. **Provide evidence** for every addressed detail: quote the exact short
   passage from the reasoning_trace that demonstrates the connection.

4. **Write a brief overall analysis** summarising how fully the hypothesis
   accounts for the complete observation.

## Grading criteria

- **Addressed (True)**: The trace contains a direct explanation of *why* or
  *how* the chosen hypothesis accounts for this specific detail.
- **Not addressed (False)**: The detail is present in the observation but the
  trace either ignores it, only restates it, or treats it as irrelevant without
  justification.

## Important rules

- Be exhaustive: do not skip minor or background details.
- Do not reward vague gestures.
- A hypothesis that explains the main event but ignores supporting details
  should receive a low coverage score.
- Base every judgement solely on what is written in the reasoning_trace --
  do not infer or assume anything that is not stated.

## Dataset-specific note (current dataset only)

{dataset_specific_note}

## Extraction rules

- **Enforce Atomicity:** Break down compound sentences and lists into atomic
  (single, indivisible) facts.
  - Example: "headache and vomiting" must be split into two separate details:
    "headache" and "vomiting".
  - Example: "Kernig and Brudzinski signs are present" must be split into
    "Kernig sign present" and "Brudzinski sign present".
  - Example: Separate every single medication, vital sign, and lab value into
    its own item.
- Extract each atomic observation fact as one item in `observation_details`.
- Use `detail` for the observation fact text, `addressed` for explicit
  linkage status, and `evidence` for a supporting quote.
- Set `addressed` to true only when the reasoning_trace explicitly links
  the detail to the chosen hypothesis.
- If `addressed` is false, leave `evidence` as an empty string.
- Be exhaustive across main and peripheral details.

## JSON output format

Return ONLY valid JSON with this structure:
{
  "overall_analysis": "Brief analysis of observation coverage in this
    reasoning_trace",
  "observation_details": [
    {
      "detail": "Specific atomic detail from the observation",
      "addressed": true,
      "evidence": "Quote from reasoning_trace showing how this detail is
        explained"
    }
  ]
}
\end{lstlisting}

\vspace{0.8em}
\noindent\textbf{User Prompt Template:}
\begin{lstlisting}[breaklines=true, basicstyle=\small\ttfamily]
Dataset: {dataset}

Analyse the following observation and reasoning_trace and produce the structured
observation-coverage evaluation.

<observation>
{full_input}
</observation>

<reasoning_trace>
{text}
</reasoning_trace>
\end{lstlisting}

\subsubsection{Evidence--Explanation Directionality (Score-Based)}
\label{app:prompt_directionality}

\noindent\textbf{System Prompt:}
\begin{lstlisting}[breaklines=true, basicstyle=\small\ttfamily]
You are an expert evaluator assessing the logical directionality of reasoning
chains generated by AI models.

## Metric Definition

**Evidence-Explanation Directionality:** The model demonstrates
awareness that abduction runs strictly from evidence to explanation (not
explanation to evidence, as in prediction/deduction).

In proper abductive reasoning, the model MUST start with the known observations
(the evidence) and logically reason forward to determine the best hypothesis
(the explanation). Reversing this logical connection -- assuming an explanation
is true and then deductively checking if it predicts or aligns with the given
observations -- is a core logical flaw for this evaluation.

## Scoring Rubric

### Score 1.0: Proper Abductive Direction (Evidence -> Explanation)
The reasoning chain explicitly treats the evidence/observations as the
foundational starting point.
- The model looks at the provided facts and constructively asks "what best
  explains this?" or "where does this evidence lead?"
- It builds its argument upward from the evidence to reach the explanatory
  conclusion.
- It never presupposes the correctness of an explanation before analyzing
  the evidence.
- Language cues: "Given these observations...", "Because we see X, it suggests
  Y...", "The evidence points toward..."

### Score 0.5: Ambiguous or Mixed Directionality
The reasoning connects evidence and explanation, but the logical flow is
fundamentally unclear or loosely structured.
- The model might list facts and the explanation side-by-side without linking
  them directionally.
- It might waver backwards and forwards, showing signs of evidence->explanation
  logic mixed tightly with explanation->evidence logic.
- The flow lacks the robust forward mathematical momentum of true abduction
  but isn't explicitly deductive either.

### Score 0.0: Backward / Deductive Direction (Explanation -> Evidence)
The reasoning chain completely reverses the abductive flow, working backward
from an assumed conclusion to the evidence.
- The model treats a candidate hypothesis as a given, explicitly deducing what
  observations *would* follow, and then checks if the evidence matches.
- It operates using a "prediction" format: "If [Hypothesis] is true, we would
  expect to see [Observation]. We see [Observation], therefore it is true."
- Any reasoning that merely justifies a pre-selected answer via post-hoc
  confirmation bias falls here.

## Dataset-Specific Guidelines

{dataset_specific_note}

## Output Format

Return exactly and ONLY a well-formatted JSON object containing the exact
fields below:
{
  "reasoning_analysis": "A concise (1-2 sentence) explanation of the logical
    directionality observed in the chain.",
  "directionality_score": 1.0
}

Note: `directionality_score` must strictly be 0.0, 0.5, or 1.0. Do not output
markdown code blocks.
\end{lstlisting}

\vspace{0.8em}
\noindent\textbf{User Prompt Template:}
\begin{lstlisting}[breaklines=true, basicstyle=\small\ttfamily]
Dataset: {dataset}

Evaluate the reasoning chain below for Evidence-Explanation Directionality
and assign a score of 0.0, 0.5, or 1.0.

**Observations / Evidence:**
<observations>
{full_input}
</observations>

**Model's Reasoning Chain:**
<reasoning_chain>
{text}
</reasoning_chain>

Assign a directionality score (0.0 / 0.5 / 1.0) and briefly explain your
assessment in the `reasoning_analysis` field.
\end{lstlisting}

\subsection{Detailed Per-Dataset Results}
\label{app:detailed_results}

Table~\ref{tab:model_results} reports per-dataset process-level scores for
DeepSeek-R1-Distill-Qwen-7B.  For counting metrics, values are mean occurrence
counts per trace; for Evidence Coverage and Directionality, values are mean
scores.  Values in bold are the maximum for that metric-dataset cell across the
three conditions.

\begin{table*}[ht]
\centering
\small
\begin{tabular}{p{3.2cm} l rrrrrrrrrr}
\toprule
& & \multicolumn{10}{c}{\textbf{Datasets}} \\
\cmidrule(lr){3-12}
& & \tabrot{ART} & \tabrot{B-COPA} & \tabrot{GoEmo.} & \tabrot{MedQA}
& \tabrot{DefNLI} & \tabrot{MuSR-M} & \tabrot{MuSR-O} & \tabrot{MuSR-T}
& \tabrot{NeuLR} & \tabrot{StratQA} \\
\textbf{Condition} & \textbf{Metric} & & & & & & & & & & \\
\midrule

\multirow[t]{7}{3.2cm}{Cor-GRPO}
& Backtracking    & 0.5 & \textbf{0.9} & 0.2 & \textbf{0.8} & 1.5
                 & 0.8 & 1.6 & 0.6 & 1.5 & 0.9 \\
& Branchiness     & 1.2 & 0.4 & 0.9 & 1.5 & 1.6
                 & 1.2 & 1.3 & 1.5 & 1.4 & 0.6 \\
& Coverage (\%)   & 43.0 & 41.0 & 23.0 & 34.0 & 46.0
                 & 49.0 & 39.0 & 44.0 & 35.0 & \textbf{37.0} \\
& Diff.\ Elim.    & 1.0 & 1.0 & \textbf{0.6} & 0.7 & 1.0
                 & 0.9 & 1.1 & 0.7 & 1.5 & 1.2 \\
& Directionality  & 0.2 & 0.1 & 0.3 & 0.1 & 0.1
                 & 0.2 & 0.1 & 0.2 & 0.2 & 0.1 \\
& Prior           & 0.4 & \textbf{0.7} & 0.1 & 0.3 & 0.5
                 & 0.9 & 0.7 & 0.3 & 1.0 & 0.4 \\
& Unc.\ Markers   & 0.6 & 0.8 & 1.0 & \textbf{1.0} & 1.1
                 & \textbf{1.3} & 0.4 & 0.8 & 1.1 & \textbf{1.1} \\
\midrule

\multirow[t]{7}{3.2cm}{CEDAR-GRPO}
& Backtracking    & \textbf{0.6} & \textbf{0.9} & 0.2 & \textbf{0.8}
                 & \textbf{1.7} & \textbf{0.9} & \textbf{1.9} & 0.6
                 & \textbf{2.3} & \textbf{1.0} \\
& Branchiness     & \textbf{1.3} & 0.5 & 1.0 & \textbf{2.8} & \textbf{2.5}
                 & \textbf{1.5} & \textbf{1.8} & \textbf{1.6} & \textbf{1.6}
                 & \textbf{0.7} \\
& Coverage (\%)   & \textbf{44.0} & \textbf{47.0} & \textbf{61.0}
                 & \textbf{48.0} & \textbf{71.0} & \textbf{52.0}
                 & \textbf{70.0} & \textbf{48.0} & \textbf{55.0} & 33.0 \\
& Diff.\ Elim.    & \textbf{1.8} & \textbf{1.1} & \textbf{0.6}
                 & \textbf{1.3} & \textbf{1.1} & \textbf{1.5}
                 & \textbf{1.3} & \textbf{1.2} & \textbf{1.7}
                 & \textbf{1.3} \\
& Directionality  & \textbf{0.71} & \textbf{0.56} & \textbf{0.49}
                 & \textbf{0.70} & \textbf{0.44} & \textbf{0.39}
                 & \textbf{0.70} & \textbf{0.71} & \textbf{0.58}
                 & \textbf{0.74} \\
& Prior           & \textbf{0.7} & 0.4 & \textbf{0.4} & \textbf{0.9}
                 & 0.5 & \textbf{1.0} & \textbf{0.9} & \textbf{0.8}
                 & \textbf{1.1} & \textbf{0.5} \\
& Unc.\ Markers   & \textbf{1.6} & \textbf{1.0} & \textbf{1.5}
                 & \textbf{1.1} & \textbf{1.5} & \textbf{1.3}
                 & \textbf{2.1} & \textbf{1.4} & 1.1 & \textbf{1.1} \\
\midrule

\multirow[t]{7}{3.2cm}{Base}
& Backtracking    & \textbf{0.6} & 0.4 & \textbf{0.5} & \textbf{0.8}
                 & 0.7 & \textbf{0.9} & 0.8 & \textbf{0.7} & 1.0 & 0.5 \\
& Branchiness     & 1.2 & \textbf{0.9} & \textbf{1.1} & 1.4 & 1.3
                 & \textbf{1.5} & 1.4 & 1.3 & \textbf{1.6} & 0.5 \\
& Coverage (\%)   & 38.0 & 22.0 & 40.0 & 25.0 & 47.0
                 & 34.0 & 45.0 & 36.0 & 23.0 & 21.0 \\
& Diff.\ Elim.    & 0.7 & 0.5 & \textbf{0.6} & 0.9 & 0.8
                 & 1.0 & 0.9 & 0.8 & 1.1 & 0.6 \\
& Directionality  & 0.1 & 0.3 & 0.2 & 0.2 & 0.2
                 & 0.2 & 0.1 & 0.1 & 0.2 & 0.5 \\
& Prior           & 0.5 & 0.3 & \textbf{0.4} & 0.7 & \textbf{0.6}
                 & 0.8 & 0.7 & 0.6 & 0.9 & 0.4 \\
& Unc.\ Markers   & 0.8 & 0.5 & 0.7 & \textbf{1.0} & 0.9
                 & 1.1 & 1.0 & 0.9 & \textbf{1.2} & 0.6 \\
\bottomrule
\end{tabular}
\caption{Process-level metric scores for DeepSeek-R1-Distill-Qwen-7B.  Values are
bolded if they are the maximum for that specific metric and dataset across all
three conditions.}
\label{tab:model_results}
\end{table*}




\begin{table*}[p]
\centering
\footnotesize
\renewcommand{\arraystretch}{1.15}
\setlength{\tabcolsep}{2.2pt}
\begin{tabularx}{\textwidth}{%
  >{\raggedright\arraybackslash}p{1.1cm}
  >{\raggedright\arraybackslash}X
  >{\raggedright\arraybackslash}X
  >{\raggedright\arraybackslash}X
  >{\raggedright\arraybackslash}X
  >{\raggedright\arraybackslash}X
  >{\raggedright\arraybackslash}X
  >{\raggedright\arraybackslash}X}
\toprule
\textbf{Dataset}
  & \textbf{Unc.\ Markers}
  & \textbf{Branch.}
  & \textbf{Back.}
  & \textbf{Diff.\ Elim.}
  & \textbf{Prior}
  & \textbf{Obs.\ Cov.}
  & \textbf{Direct.} \\
\midrule
ART
  & Focus on uncertainty while reasoning about which hypothesis better explains observations.
  & Do not count simple option selection; count only internal exploration.
  & Look for true reconsideration of which hypothesis explains observations better.
  & Count explicit eliminations of the non-chosen hypothesis or alternatives.
  & Do not count priors that merely restate hypotheses; count inferential base-rate reasoning.
  & Treat `Observation 1' and `Observation 2' as complete; extract all details.
  & Check reasoning starts from observations and seeks hypothesis, not back-fitting. \\
\addlinespace
B-COPA
  & Focus on uncertainty in causal reasoning and option comparison.
  & Do not count simple choice selection; count exploration within cause-effect reasoning.
  & Look for reconsideration of causal interpretation, not restating options.
  & Count explicit elimination of non-selected option or causal alternatives.
  & Priors about typical everyday cause-effect relationships.
  & Treat `Cause:' as observation; extract relevant details.
  & `Cause' is evidence; model must evaluate which option is most plausible effect. \\
\addlinespace
DefNLI
  & Uncertainty indicating defeasible or non-certain inference relations.
  & Exploration of stronger vs.\ weaker inferences and defeasible conclusions.
  & Reconsideration when inference appears defeated or less robust.
  & Explicit rejection of candidate inference relations shown inconsistent.
  & Priors about typical premise-hypothesis relations and defaults.
  & Treat `Premise', `Hypothesis', `Update' together as complete observation.
  & Reasoning evaluates logical impact of Update from given text. \\
\addlinespace
GoEmo.
  & Uncertainty in label selection; avoid counting task-domain terms.
  & Count only if model explores multiple emotion labels.
  & Count only genuine revisions in emotion-label selection.
  & Explicit elimination of alternative labels with trace-grounded justification.
  & Priors about emotions typically associated with contexts.
  & Treat `Text' as complete observation; extract salient spans.
  & Reasoning extracts cues before selecting emotion label. \\
\bottomrule
\end{tabularx}
\caption{Dataset-specific notes for all process-level metrics injected into the judge prompt. Each cell gives the exact text appended to the shared system prompt for that dataset.}
\label{tab:notes_all_metrics}
\end{table*}

\begin{table*}[p]
\centering
\footnotesize
\renewcommand{\arraystretch}{1.15}
\setlength{\tabcolsep}{2.2pt}
\begin{tabularx}{\textwidth}{%
  >{\raggedright\arraybackslash}p{1.1cm}
  >{\raggedright\arraybackslash}X
  >{\raggedright\arraybackslash}X
  >{\raggedright\arraybackslash}X
  >{\raggedright\arraybackslash}X
  >{\raggedright\arraybackslash}X
  >{\raggedright\arraybackslash}X
  >{\raggedright\arraybackslash}X}
\toprule
\textbf{Dataset}
  & \textbf{Unc.\ Markers}
  & \textbf{Branch.}
  & \textbf{Back.}
  & \textbf{Diff.\ Elim.}
  & \textbf{Prior}
  & \textbf{Obs.\ Cov.}
  & \textbf{Direct.} \\
\midrule
MuSR
  & Uncertainty in narrative interpretation and conclusion drawing.
  & Exploration of different narrative interpretations or solutions.
  & Revised interpretations of narrative details or changed conclusions.
  & Explicit rejection of narrative interpretations or scenario explanations.
  & Priors about typical narrative behavior, motives, patterns.
  & Treat `Context' and `Problem' as complete; extract narrative details (actors, timing, locations).
  & Reasoning builds explanations from details, not assumes conclusion and back-fits. \\
\addlinespace
MedQA
  & Extract markers only from model's own reasoning, not probabilistic symptoms.
  & Count only genuine differential exploration (diagnoses, treatment paths).
  & Corrections in diagnostic/treatment reasoning or missed details.
  & Each answer choice explicitly ruled out with clinical rationale.
  & Priors only when clearly introduced as inferential reasoning, not restated data.
  & Treat `Problem' as observation; extract all clinical details.
  & Reasoning starts from clinical evidence and moves toward diagnosis/treatment. \\
\addlinespace
NeuLR
  & Uncertainty while weighing competing abductive explanations.
  & Do not count superficial option comparison; genuine exploration.
  & True reconsideration of which explanation best fits observations.
  & Explicit elimination moves using concrete mismatch evidence.
  & Priors justifying why one hypothesis more plausible.
  & Treat `Logical Rules and Known Facts' and `Target Conclusion' as complete.
  & Reasoning uses Rules/Facts to find Missing Fact, not work backwards. \\
\addlinespace
StratQA
  & Uncertainty in multi-step reasoning and evidence-to-conclusion transitions.
  & Exploration of multiple reasoning paths or inference chains.
  & Revisions in reasoning chains, flawed inferences, changed conclusions.
  & Explicit rejection of competing YES/NO reasoning lines.
  & World-knowledge priors and typicality assumptions.
  & Treat `Question' and `Evidence' as complete; extract relevant facts.
  & Reasoning builds upon facts toward YES/NO answer, not cherry-picks post-hoc. \\
\bottomrule
\end{tabularx}
\addtocounter{table}{-1}
\caption{Dataset-specific notes for all process-level metrics injected into the judge prompt. Each cell gives the exact text appended to the shared system prompt for that dataset. (continued)}
\end{table*}

\FloatBarrier
\section{SFT Data Construction and Training Details}
\label{app:sft_dataset_creation}

To compare CEDAR-GRPO against supervised fine-tuning, we construct an SFT dataset from exactly the same training sources used in the GRPO setting. The original examples contain task inputs and gold final answers, but they do not consistently provide a complete rationale in the output format used during RL training. We therefore generate synthetic reference rationales and insert them into the model response as the content of the \texttt{<think>} block, while keeping the original gold label, sentence set, missing fact, or program as the \texttt{<answer>} target.

Synthetic rationales are generated with Gemini 3.0 Flash Preview through the OpenRouter chat-completions API. For all datasets, we use temperature $0.2$, top-$p=0.9$, up to six retries per example, eight concurrent API workers, and a 90-second request timeout. The maximum completion budget is 350 tokens for UniADILR, Balanced COPA, CLIMATE-FEVER, and AbductionRules; 450 tokens for CauseLogics; 900 tokens for Crypto; and 1000 tokens for List Function. The generator is given the problem input together with the gold answer and is instructed to produce a concise rationale that justifies the gold output. The resulting SFT target has the same response structure as the GRPO policy outputs:

\begin{lstlisting}[breaklines=true, basicstyle=\small\ttfamily]
<think>
{synthetic_rationale}
</think>
<answer>
{gold_answer}
</answer>
\end{lstlisting}

We apply strict validation before accepting a generated rationale. Each completion must exactly match the required \texttt{<think>}--\texttt{<answer>} format, and the answer block must match the known gold answer. For classification and selection datasets, this means exact matching against the normalized label or sentence-number set. For AbductionRules, the answer must exactly match the gold missing fact. For the two code-generation datasets, the generated answer is also compiled and executed against the held-out test cases for that example; the rationale is accepted only if the generated \texttt{transform} function passes all tests. The accepted dataset stores only the validated rationale as an added field, leaving the original gold answer or gold function as the final supervised answer. Failed or malformed generations are retried; unresolved failures are left without a rationale and can be resumed later.

The final mixed SFT dataset uses the same train/validation split as the GRPO data: 1,920 training examples and 480 validation examples. UniADILR, Balanced COPA, CauseLogics, CLIMATE-FEVER, and AbductionRules each contribute 320 training and 80 validation examples; Crypto and List Function each contribute 160 training and 40 validation examples. Each sample is additionally tagged with its dataset name before the mixed split is shuffled with seed 42.

\subsection{SFT optimization and model selection}
\label{app:sft_training_setup}

We train the Qwen3-4B and DeepSeek-R1-Distill-Qwen-7B SFT baselines with TRL's \texttt{SFTTrainer} on the full mixed split described above. Each example uses the same dataset-specific system and user prompts as in GRPO, followed by the validated \texttt{<think>}--\texttt{<answer>} target. The cross-entropy loss is applied only to assistant tokens; system and user tokens are masked.

\begin{table}[t]
\centering
\footnotesize
\setlength{\tabcolsep}{3.5pt}
\renewcommand{\arraystretch}{1.06}
\begin{tabularx}{\columnwidth}{@{}lX@{}}
\toprule
\textbf{Setting} & \textbf{SFT configuration} \\
\midrule
Data & 1,920 training and 480 validation examples; sequence packing disabled \\
Quantization / LoRA & 4-bit NF4; rank $r=64$; $\alpha=64$; Unsloth gradient checkpointing \\
LoRA targets & \texttt{q\_proj}, \texttt{k\_proj}, \texttt{v\_proj}, \texttt{o\_proj}, \texttt{gate\_proj}, \texttt{up\_proj}, \texttt{down\_proj} \\
Sequence length & 4,096 tokens \\
Optimizer & AdamW (Torch), learning rate $1\times10^{-5}$, $(\beta_1,\beta_2)=(0.9,0.99)$, weight decay $0.1$ \\
Schedule / clipping & Cosine decay, 2 warmup steps, maximum gradient norm $0.1$ \\
Batch / budget & Single-GPU train/eval batch size 4; gradient accumulation 1 (effective train batch size 4); five epoch (2400 optimizer steps) \\
Checkpointing & Adapter checkpoints every 128 optimizer steps \\
Validation & All 480 examples every 128 steps and at epoch end; one sampled completion per example ($T=0.7$, top-$p=0.95$, maximum 2,048 new tokens) \\
Selection & Saved checkpoint with the highest mean validation correctness under the dataset-specific exact-, set-, or execution-based verifier \\
\bottomrule
\end{tabularx}
\caption{Optimization and model-selection setup for the SFT ablation. Held-out evaluation tasks are not used for checkpoint selection.}
\label{tab:sft_training_setup}
\end{table}
\FloatBarrier

\subsection{Synthetic-rationale prompt templates}

The prompts below are the dataset-specific templates used to generate the SFT rationales. In each case, the generator receives the original task input and the gold answer, and must return only a formatted SFT target.

\subsubsection{Prompt Template: UniADILR}

\noindent\textbf{System Prompt:}
\begin{lstlisting}[breaklines=true, basicstyle=\small\ttfamily]
You are generating supervised fine-tuning data for an abductive evidence-selection model.

You will receive:
1. A context with numbered sentences
2. A hypothesis
3. The gold supporting sentence numbers

Write a concise rationale explaining why the gold sentences provide the necessary evidence for the hypothesis.

The rationale should:
- Start from the hypothesis.
- Identify what needs to be explained or supported.
- Explain how each gold sentence contributes to the explanation.
- Explain why the selected sentences work together to support the hypothesis.
- Ignore unrelated context sentences.
- Avoid external knowledge and unsupported speculation.

Output exactly:

<think>
[A concise abductive evidence-selection rationale, usually 3-5 sentences.]
</think>

<answer>
[The gold sentence numbers only, comma-separated.]
</answer>

Do not change the gold sentence numbers in the answer section.
Do not include the word "sent" in the answer section.
Do not include any text, punctuation, or explanation other than comma-separated numbers in the answer section.
\end{lstlisting}

\noindent\textbf{User Prompt:}
\begin{lstlisting}[breaklines=true, basicstyle=\small\ttfamily]
Context:
{context_str}

Hypothesis:
{hypothesis}

Gold Supporting Sentences:
{gold_answer}

Generate the SFT target.
\end{lstlisting}

\subsubsection{Prompt Template: Balanced COPA}

\noindent\textbf{System Prompt:}
\begin{lstlisting}[breaklines=true, basicstyle=\small\ttfamily]
You are generating supervised fine-tuning data for an abductive causal reasoning model.

You will receive:
1. A premise describing a situation or event
2. Two possible causes
3. The gold answer

Write a concise rationale explaining why the gold answer is the more plausible cause of the premise.

The rationale should:
- Start from the premise.
- Evaluate both choices as possible causes.
- Explain why the gold choice more directly or plausibly leads to the premise.
- Explain why the other choice is weaker, less direct, or less plausible.
- Use common sense and typical real-world causal relationships.
- Avoid unrelated speculation.

Output exactly:

<think>
[A concise abductive causal rationale, usually 3-5 sentences.]
</think>

<answer>
[The gold answer only: either 1 or 2.]
</answer>

Do not change the gold answer in the answer section.
Do not include any text, punctuation, or explanation in the answer section.
\end{lstlisting}

\noindent\textbf{User Prompt:}
\begin{lstlisting}[breaklines=true, basicstyle=\small\ttfamily]
Premise: {premise}

Choice 1: {choice1}
Choice 2: {choice2}

Gold Answer:
{gold_answer}

Generate the SFT target.
\end{lstlisting}

\subsubsection{Prompt Template: CauseLogics}

\noindent\textbf{System Prompt:}
\begin{lstlisting}[breaklines=true, basicstyle=\small\ttfamily]
You are generating supervised fine-tuning data for an abductive logical reasoning model.

You will receive:
1. A set of premises
2. A set of rules
3. An observed phenomenon
4. A possible cause
5. The gold answer

Write a concise rationale explaining why adding the possible cause does or does not make the phenomenon logically inferable.

The rationale should:
- Start from the phenomenon.
- Assume the possible cause is added to the premises.
- Identify the relevant rule or rule chain.
- State which required facts are already present or can be derived.
- Explain whether forward reasoning can infer the phenomenon.
- If the phenomenon cannot be inferred, briefly state where the proof path fails.
- Use only the given premises, rules, and possible cause.
- Avoid unrelated facts, unrelated rules, and external knowledge.

Output exactly:

<think>
[A concise abductive logical rationale, usually 3-6 sentences.]
</think>

<answer>
[The gold answer only: TRUE or FALSE.]
</answer>

Do not change the gold answer in the answer section.
Do not include any text, punctuation, or explanation in the answer section.
\end{lstlisting}

\noindent\textbf{User Prompt:}
\begin{lstlisting}[breaklines=true, basicstyle=\small\ttfamily]
Premises:
{premises_text}

Rules:
{rules_text}

Phenomenon:
{phenomenon}

Possible Cause:
{possible_cause}

Gold Answer:
{gold_answer}

Generate the SFT target.
\end{lstlisting}

\subsubsection{Prompt Template: CLIMATE-FEVER}

\noindent\textbf{System Prompt:}
\begin{lstlisting}[breaklines=true, basicstyle=\small\ttfamily]
You are generating supervised fine-tuning data for an evidence-grounded fact-checking model.

You will receive:
1. A claim
2. A list of evidences
3. The gold answer

Write a concise rationale explaining why the evidence supports, refutes, is insufficient to evaluate, or disputes the claim.

The rationale should:
- Start from the claim.
- Identify the most relevant evidence.
- Explain how the evidence relates to the claim.
- For SUPPORTS, explain why the evidence makes the claim more likely true.
- For REFUTES, explain why the evidence contradicts the claim.
- For NOT ENOUGH INFO, explain what key information is missing.
- For DISPUTED, explain that the provided evidence is mixed, conflicting, or does not lead to a single clear verdict.
- Use only the provided evidence, not external knowledge.
- Avoid unrelated evidence and unsupported speculation.

Output exactly:

<think>
[A concise evidence-grounded rationale, usually 3-5 sentences.]
</think>

<answer>
[The gold answer only: SUPPORTS, REFUTES, NOT ENOUGH INFO, or DISPUTED.]
</answer>

Do not change the gold answer in the answer section.
Do not include any text, punctuation, or explanation in the answer section.
\end{lstlisting}

\noindent\textbf{User Prompt:}
\begin{lstlisting}[breaklines=true, basicstyle=\small\ttfamily]
Claim:
{claim}

Evidence:
{evidence_text}

Gold Answer:
{gold_answer}

Generate the SFT target.
\end{lstlisting}

\subsubsection{Prompt Template: AbductionRules}

\noindent\textbf{System Prompt:}
\begin{lstlisting}[breaklines=true, basicstyle=\small\ttfamily]
You are generating supervised fine-tuning data for an abductive reasoning model.

You will receive:
1. A context with facts and rules
2. A query
3. The gold missing fact

Write a concise rationale explaining why the gold missing fact makes the query decidable from the context.

The rationale should:
- Start from the query.
- Identify the relevant rule or rule chain.
- State which required facts are already present.
- State what the gold missing fact adds.
- Explain whether the query becomes provable or disprovable.
- Ignore unrelated facts and rules.

Output exactly:

<think>
[A concise abductive rationale, usually 3-5 sentences.]
</think>

<answer>
[The gold missing fact only, as one sentence ending with a period.]
</answer>

Do not change the gold missing fact in the answer section.
Do not include more than one fact in the answer section.
\end{lstlisting}

\noindent\textbf{User Prompt:}
\begin{lstlisting}[breaklines=true, basicstyle=\small\ttfamily]
Context:
{context}

Query:
{query}

Gold Missing Fact:
{gold_answer}

Generate the SFT target.
\end{lstlisting}

\subsubsection{Prompt Template: Crypto}

\noindent\textbf{System Prompt:}
\begin{lstlisting}[breaklines=true, basicstyle=\small\ttfamily]
You are generating supervised fine-tuning data for a rule-induction-to-code model.

You will receive:
1. Several training examples with input and output strings
2. A gold Python function that implements the intended hidden transformation

Write a concise rationale explaining why the gold transformation fits the training examples, then provide a Python implementation of the same rule.

The rationale should:
- Start from the input-output examples.
- Identify the hidden character-level string transformation.
- Explain how the rule accounts for the outputs across examples.
- Mention whether the rule is a Caesar shift, Atbash mapping, or another exact character mapping when relevant.
- Briefly rule out a simpler wrong pattern if helpful.
- Emphasize that the rule should generalize beyond the shown examples.
- Avoid unrelated speculation.

Output exactly:

<think>
[A concise rule-induction rationale, usually 3-5 sentences.]
</think>

<answer>
[Python code only.]
</answer>

Code requirements:
- Define exactly one function named transform.
- The function takes one argument: s.
- The function must return a string.
- Use the gold function only to infer the intended rule.
- Preserve the behavior implied by the examples for lowercase letters, uppercase letters, and any non-letter characters.
- Do not hardcode the training examples.
- Do not use imports, printing, input(), or randomness.
- Do not include markdown code blocks.
- Do not include any text outside the Python function in the answer section.
\end{lstlisting}

\noindent\textbf{User Prompt:}
\begin{lstlisting}[breaklines=true, basicstyle=\small\ttfamily]
Training examples:
{train_prompt}
{split_text}

Gold Function:
{gold_function}

Generate the SFT target.
\end{lstlisting}

For Crypto, at most the first ten normal training examples are included in the rationale-generation prompt. When available, the split metadata is included as a transformation-type hint. The generated code is accepted only if it passes the example's held-out test cases.

\subsubsection{Prompt Template: List Function}

\noindent\textbf{System Prompt:}
\begin{lstlisting}[breaklines=true, basicstyle=\small\ttfamily]
You are generating supervised fine-tuning data for a rule-induction-to-code model.

You will receive:
1. Several training examples with input and output lists
2. A gold Python function that implements the intended hidden transformation

Write a concise rationale explaining why the gold transformation fits the training examples, then provide a Python implementation of the same rule.

The rationale should:
- Start from the input-output examples.
- Identify the hidden list transformation.
- Explain how the rule accounts for the outputs.
- Briefly rule out a simpler wrong pattern if helpful.
- Emphasize that the rule should generalize beyond the shown examples.
- Avoid unrelated speculation.

Output exactly:

<think>
[A concise rule-induction rationale, usually 3-5 sentences.]
</think>

<answer>
[Python code only.]
</answer>

Code requirements:
- Define exactly one function named transform.
- The function takes one argument: lst.
- The function must return a list of integers.
- Use the gold function only to infer the intended rule; adapt the function name and argument name to transform(lst).
- Do not hardcode the training examples.
- Do not use imports, printing, input(), or randomness.
- Do not include markdown code blocks.
- Do not include any text outside the Python function in the answer section.
\end{lstlisting}

\noindent\textbf{User Prompt:}
\begin{lstlisting}[breaklines=true, basicstyle=\small\ttfamily]
Training examples:
{train_prompt}

Gold Function:
{gold_function}

Generate the SFT target.
\end{lstlisting}

For List Function, the generated code must define a single \texttt{transform(lst)} function and pass all held-out test cases for that example before the accompanying rationale is accepted.

\section{Additional Dataset Notes for the General-Reasoning Ablation}
\label{app:general_reasoning_data}

For the general-reasoning ablation, we construct a balanced auxiliary training mixture from six widely used reasoning benchmarks: GSM8K, FOLIO, CommonsenseQA, VitaminC, MMLU, and BIG-Bench~\citep{cobbe2021gsm8k,han2024folio,talmor2019commonsenseqa,schuster2021vitaminc,hendrycks2021mmlu,srivastava2022bigbench}. The goal of this ablation is not to optimize for any single benchmark family, but to test whether improvements attributed to abductive training persist when the auxiliary signal is replaced by a broader mixture of mathematical, logical, commonsense, fact-verification, and academic question-answering tasks. To keep the comparison controlled, each source contributes the same number of instances. The resulting corpus contains 2{,}400 examples in total, split into 1{,}920 training examples and 480 validation examples.

\subsection{Mixture composition}

\begin{table*}[h!]
\centering
{\footnotesize
\setlength{\tabcolsep}{3.5pt}
\renewcommand{\arraystretch}{1.08}
\begin{tabular}{p{3.2cm}p{2.6cm}cp{6.3cm}}
\toprule
Dataset & Reasoning profile & Samples & Role in the ablation mixture \\
\midrule
\makecell[l]{GSM8K\\\citep{cobbe2021gsm8k}}
&
\makecell[l]{Arithmetic,\\multi-step reasoning}
& 400
&
Introduces numerical and procedural reasoning through grade-school math word problems. Final supervision is based on the extracted numerical answer. \\
\makecell[l]{FOLIO\\\citep{han2024folio}}
&
\makecell[l]{Formal logical\\inference}
& 400
&
Provides premise--conclusion judgments grounded in first-order logical structure, adding a strongly symbolic component to the mixture. \\
\makecell[l]{CommonsenseQA\\\citep{talmor2019commonsenseqa}}
&
\makecell[l]{Commonsense\\multiple choice}
& 400
&
Supplies broad everyday reasoning questions in a fixed five-choice format, complementing the more formal logical tasks. \\
\makecell[l]{VitaminC\\\citep{schuster2021vitaminc}}
&
\makecell[l]{Fact verification\\under contrastive evidence}
& 400
&
Adds evidence-sensitive classification in which small textual changes can reverse the correct label, making it useful for testing fine-grained judgment. \\
\makecell[l]{MMLU\\\citep{hendrycks2021mmlu}}
&
\makecell[l]{Academic\\multiple choice}
& 400
&
Contributes subject-diverse question answering from a reasoning-oriented subset of academic domains. \\
\makecell[l]{BIG-Bench\\\citep{srivastava2022bigbench}}
&
\makecell[l]{Heterogeneous\\multiple-choice reasoning}
& 400
&
Adds task diversity through four reasoning-focused multiple-choice tasks spanning deduction, state tracking, date reasoning, and tabular reasoning. \\
\bottomrule
\end{tabular}
\caption{
Composition of the general-reasoning ablation mixture. Each source contributes 400 instances, producing a balanced 2{,}400-example corpus with an 80/20 train/validation split.
}
\label{tab:general_reasoning_data}
}
\end{table*}

\subsection{Dataset-specific notes}

\paragraph{GSM8K.}
GSM8K is a benchmark of grade-school mathematics problems intended to test multi-step quantitative reasoning~\citep{cobbe2021gsm8k}. We sample 400 examples from the public training data and partition them into 320 training and 80 validation instances. Although the benchmark includes worked solutions, the supervision target in this ablation is the final numeric answer, so this component primarily evaluates whether general reasoning fine-tuning improves reliable answer derivation rather than free-form explanation quality.

\paragraph{FOLIO.}
FOLIO is a natural-language reasoning benchmark with first-order-logic structure, where a model must determine whether a conclusion follows from a set of premises~\citep{han2024folio}. We use 320 training and 80 validation examples drawn from the released train and validation portions. Since the original benchmark includes an uncertainty class, this component is especially useful for probing whether the model can preserve calibrated three-way logical judgments under mixed-task training.

\paragraph{CommonsenseQA.}
CommonsenseQA is a five-way multiple-choice benchmark designed to require everyday background knowledge rather than shallow textual matching~\citep{talmor2019commonsenseqa}. We include 320 training and 80 validation questions sampled from the released train and validation data. Within the ablation mixture, CommonsenseQA serves as the main source of broad commonsense supervision, counterbalancing the more formal or domain-specific benchmarks.

\paragraph{VitaminC.}
VitaminC is a fact-verification benchmark built around contrastive evidence edits, such that small revisions in wording, numbers, or negation may flip the correct label~\citep{schuster2021vitaminc}. We sample 320 training and 80 validation instances while approximately preserving the original class proportions. In the resulting subset, the training split contains 160 \texttt{SUPPORTS}, 114 \texttt{REFUTES}, and 46 \texttt{NOT ENOUGH INFO} examples, while the validation split contains 40, 29, and 11 examples respectively. This component is included to test whether the model can maintain sensitivity to fine-grained evidential distinctions under broader reasoning supervision.

\paragraph{MMLU.}
MMLU is a large multitask benchmark spanning many academic subjects and levels of expertise~\citep{hendrycks2021mmlu}. For the ablation, we use a focused eight-subject subset: abstract algebra, formal logic, logical fallacies, college computer science, high-school statistics, high-school physics, econometrics, and high-school world history. We sample 50 questions per subject and split each subject into 40 training and 10 validation examples, yielding 320 training and 80 validation instances overall. This design preserves topical breadth while biasing the mixture toward subjects with a clearer reasoning component.

\paragraph{BIG-Bench.}
BIG-Bench is a collaborative benchmark suite intended to probe a wide range of language-model capabilities~\citep{srivastava2022bigbench}. We restrict attention to four reasoning-oriented multiple-choice tasks: \texttt{logical\_deduction}, \texttt{tracking\_shuffled\_objects}, \texttt{date\_understanding}, and \texttt{penguins\_in\_a\_table}. Each task contributes 100 examples, split into 80 training and 20 validation instances. This portion of the mixture broadens task diversity beyond standard question answering by incorporating explicit deduction, state tracking, temporal reasoning, and lightweight table reasoning.

\subsection{Interpretation}

This ablation mixture is intentionally heterogeneous. GSM8K emphasizes quantitative reasoning; FOLIO emphasizes formal logical validity; CommonsenseQA emphasizes everyday inference; VitaminC emphasizes evidence-conditioned judgment; and MMLU and BIG-Bench introduce broader subject and task diversity. The equal allocation of 400 examples per source prevents the ablation from being dominated by any single benchmark family and makes the comparison against abductive training easier to interpret.

At the same time, this corpus should not be read as a benchmark in its own right. It is a controlled training mixture designed for ablation analysis. The purpose of the condition is to test whether gains arise specifically from abductive supervision, or whether a comparably sized but more general reasoning curriculum yields similar improvements.

\end{document}